\documentclass[preprint]{bmcart}

\usepackage{amsthm,amsmath}
\usepackage[utf8]{inputenc} 
\usepackage{natbib}
\usepackage{graphicx}
\usepackage{multirow}
\usepackage{tabularx} 
\usepackage{caption}
\usepackage{tikz}

\def\myacm{0}

\startlocaldefs
\endlocaldefs

\begin{document}

\begin{frontmatter}

\begin{fmbox}
\dochead{Review-PrePrint}


\title{A Survey on GAN Acceleration Using Memory Compression Techniques}


\author[
  addressref={cairo},                   
  corref={cairo},                       
  email={dina.tantawy@eng.cu.edu.eg}   
]{\inits{D.T.} \fnm{Dina} \snm{Tantawy}}

\author[
  addressref={nyu},
  email={mzahran@cs.nyu.edu}
]{\inits{M.Z.}\fnm{Mohamed} \snm{Zahran}}

\author[
  addressref={cairo},
  email={wassal@eng.cu.edu.eg}
]{\inits{A.W.}\fnm{Amr} \snm{Wassal}}


\address[id=cairo]{
  \orgdiv{Department of Computer Engineering},             
  \orgname{Cairo University},          
  \city{Cairo},                              
  \cny{Egypt}                                    
}
\address[id=nyu]{%
  \orgdiv{Courant institution of Mathematical Sciences},
  \orgname{NewYork Univeristy},
  \city{NewYork},
  \cny{USA}
}



\end{fmbox}


\begin{abstractbox}

\begin{abstract} 
Since its invention, Generative adversarial networks (GANs) have shown outstanding results in many applications. Generative Adversarial Networks are powerful yet, resource-hungry deep-learning models.  Their main difference from ordinary deep learning models is the nature of their output. For example, GAN output can be a whole image versus other models detecting objects or classifying images. Thus, the architecture and numeric precision of the network affect the quality and speed of the solution. Hence, accelerating GANs is pivotal. Accelerating GANs can be classified into three main tracks: (1) Memory compression, (2) Computation optimization, and (3) Data-flow optimization. Because data-transfer is the main source of energy usage, memory compression leads to the most savings. Thus, in this paper, we survey memory compression techniques for CNN-Based GANs.  Additionally, the paper summarizes opportunities and challenges in GANs acceleration and suggests open research problems to be further investigated.

\end{abstract}




\end{abstractbox}
%

\end{frontmatter}



\section{Introduction}

Recently, Deep learning \textbf{(DL)} applications are getting unprecedented fame. Many Machine (Deep) learning applications are used daily like face recognition, voice recognition, weather predictions, image super-resolution, \dots etc. Companies and researchers alike compete to present more applications each day and enhance existing ones. They also compete to make them affordable and usable by everyone. 

Deep Generative models  have been on a rise as well.
GAN or Generative adversarial network is one of the most famous generative models\cite{goodfellow2014generative}. It consists of at least two networks competing against each other. It has been used in many applications like speech synthesis \cite{speech_bollepalli2019generative}, text-to-image translation \cite{txttoimg_zhang2017stackgan,txttoimg_zhang2018photographic}, image-to-image translation \cite{imgtoimg_choi2018stargan}, image super resolution \cite{highres_wang2018high}, music generation \cite{music_gansynth},  videos synthesis \cite{video_gan_clark2019efficient}, \dots etc.

Having one network in deep learning is very computationally intensive, having two networks or more is even worse. Additionally, GANs training  is susceptible to divergence or mode collapses.
 GANs training  instability adds an extra layer of complexity.

Additionally, there is an increasing need for  running generators of GANs on embedded devices. For example, Using style transfer to make clothes fitting at stores \cite{app_GAN_tryon_Fwgan}. Another usage for GANs on hand-held device is using video super-resolution to save bandwidth while downloading videos  \cite{app_GAN_superresolution}. 
 Thus, current techniques and accelerators need to be revisited and adapted to serve the urgent demand of GANs.

Accelerating GANs goals are power efficiency, speed, and solution quality. In real world, it is hard to improve everything, a price must be paid according to no free-lunch theorem. Depending on the application, the importance of one goal over the other will vary and thus the optimization technique as well.
Acceleration (optimization) techniques target three main categories:
\begin{itemize}
    \item Memory compression: this category uses compression techniques to minimize memory requirement  while preserving solution quality which in turn saves  energy usage. 
    \item Computation optimization: this category uses mathematical transformation and circuit optimization to decrease the number of mathematical operations or cycles needed alongside  optimizing their needed power and increase their speed.
    \item Dataflow optimization: this category uses mapping, scheduling, and reordering data and/or operations to maximize data reuse and minimize ineffectual operations which in turn will save energy and enhance throughput.
    
\end{itemize}

One bottleneck of running GANs  is the huge cost endured by data transfer to/from accelerating-chip memory followed by the computation cost and eventually the overhead of data transfer between different units. Thus, our approach is to investigate optimization techniques tackling those areas one by one, starting by the memory compression because of its huge cost. Thus, this paper presents a survey about recent efforts made to accelerate GANs using memory compression. Our work will focus on GANs generating images although it applies to other types of GAN tasks as well. Computational optimization and dataflow optimization are considered for a future work.

Our work is complementary to other surveys papers mentioned in sec.\ref{related}, highlighting special issues facing GANs and different efforts done to resolve them. This paper has the following contributions:
\begin{itemize}
    \item To our best knowledge, this is the first paper to survey GAN compression.
    \item Providing a taxonomy for GAN optimization 
    \item Summarizing recent research work in accelerating GANs.
    \item Providing open research questions for accelerating GANs.
\end{itemize}

The remainder of this paper is organized as follows, section \ref{back} presents a brief background on how GANs work.   Section \ref{sec:MemoryMinimization} reviews different efforts to design GAN accelerators. Section \ref{related} highlights related work and main differences. Finally, Section \ref{summary} concludes the work and presents open research questions that need to be further studied.

\section{Background}\label{back}

Generative Adversarial Network (\textbf{GAN}) is a type of generative model that uses deep learning (DL) techniques to generate data. 
As mentioned earlier, GAN is more of a way of training different smaller models to compete against each other than being a newly devised model. 

CNN-based GANs can have many different architectures like DCGAN by \cite{DCGAN}, Pix2Pix by \cite{imgtoimg_isola2017image}, ...etc. Although different CNN-based GAN models have different architectures, Most of them share the same underlying concept of having competitive networks and using transpose convolution or up sampling layers. In this section, we will use DCGAN as a representative model that shares a common base with others to explain the idea of GANs in general.

\begin{figure}[th!]
    \centering
    \includegraphics[width=0.9\textwidth]{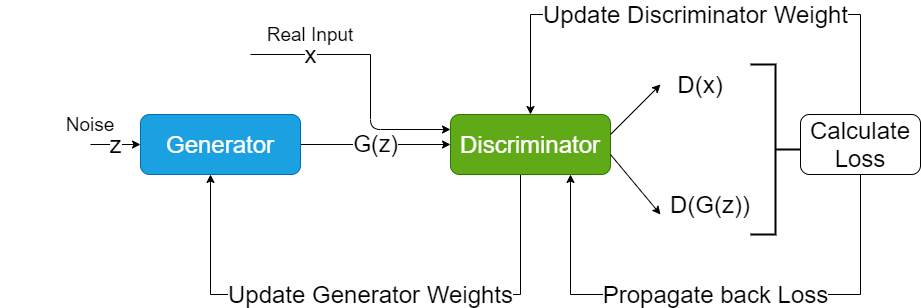} 
     \if\myacm1 
        \Description{Gan training cycles} 
     \fi
    \caption{\textbf{Example of GAN general organization}}
    \label{fig:GAN}
    \end{figure}

\textit{Structure:} GAN model consists of at least two Networks\footnote{some applications like style-transfer requires more than one GAN, more than two Networks.}: Generator and Discriminator in an organization like fig.  \ref{fig:GAN}. In training, the generator takes an input noise ($z$) generate data $G(z)$ that looks more like a real one, while the discriminator tries to be better at discriminating the generated data from the real one $x$. Thus, the goal of the discriminator is opposite to that of the generator. That's why it is called ``adversarial'' as they are competing against each other. The training ends, when the discriminator cannot enhance its accuracy anymore, and the evaluation metrics, as described later in this section, are satisfactory. Sometimes the training does not converge, and measures and limits are used to halt the training process and restart it.
 
The optimization problem can be represented as the following equation
\begin{align} \label{eq:GAN_LOSS}
    \mathcal{L}_{gan} &= min_{G}max_{D} V(D,G) = \nonumber \\ &E_{x \sim p_{data}(x) } [log D(x)]          
     + E_{z \sim p_{z}(z)} [log (1-D(G(z))]
\end{align}
where $G$ represents the generator, $D$ represents the Discriminator, $p_{data}(x)$ represents the original data distribution, and $p_{z}(z)$ represents the noise input distribution.
The first part of the equation $E_{x \sim p_{data}(x) } [log D(x)]$ represents the probability that the discriminator classifies real data as real. While second part of the equation $E_{z \sim p_{z}(z)} [log (1-D(G(z))]$ represents the ability of discriminator to classify fake data as fake. the discriminator tries to maximize those two parts of the equation, that's why we need $max_{D} V(D,G)$. On the other hand, the generator tries to fool the discriminator to detect fake images as real, thus it tries to minimize the second part of the equation and adding the $min_{G}  V(D,G)$. Combining the two network optimizations, we get the above equation.

\begin{figure}[th!]
    \centering
    \includegraphics[width=0.9\textwidth]{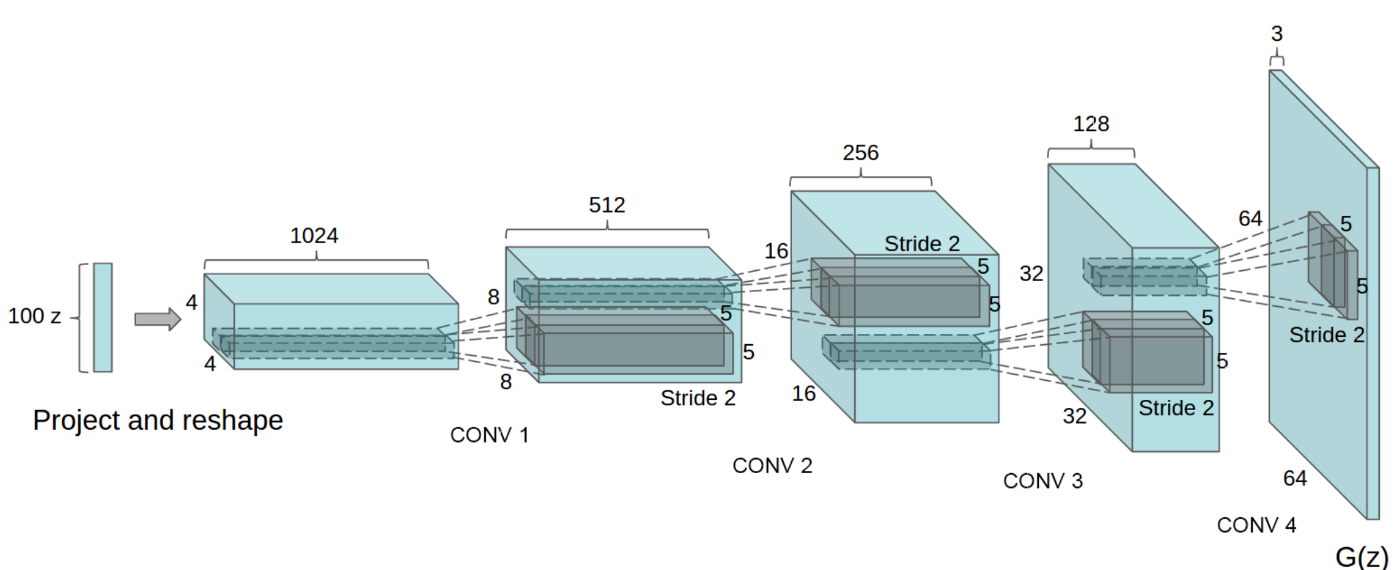}
    \if\myacm1 
    \Description{DCGAN Generator Architecture} 
    \fi
    \caption{\textbf{Generator Model in DCGAN \cite{DCGAN}}}
    \label{fig:DCGAN_G}
    \end{figure}

Discriminator Network is an ordinary CNN or LSTM or any other deep-learning classification/regression models. In contrast to discriminator that outputs only a decision or a prediction, the generator generates the data itself, be it an image, music notes, animated character...etc. Thus, the output of the generator network is larger than its input. Fig.  \ref{fig:DCGAN_G} shows the generator Network in DCGAN. Like most generator networks, the input is a noise vector or initial image that gets expanded and reshaped to a bigger size. This expansion and reshaping are performed to train the convolution to act as a Transposed Convolution.

\textit{Evaluation Metrics:} GANs have many evaluation metrics. we can split metrics into two types: 1) Functional metrics and 2) Performance metrics. Functional metrics measure how good the result of a GAN towards the target functionality. We can also name them as "quality" metrics or scores. Most commonly used function metrics are Inception Score (IS), Fréchet Inception Distance (FID), Peak-signal-to-noise-Ratio (PSNR), mean Pixel accuracy (mPA) and Intersection over Union (IoU) . Both IS and FID  measure the quality of the generated outputs, and their diversity. More information about how they are calculated can be found in the following work\cite{InceptionScore,FID}. The higher the IS the better the diversity in the generated data. While the lower the FID, the better image quality is produced. PSNR is also used especially in blending images or creating a super-resolution image (the higher the better). mPA measures the mean difference in percentage between the generated image and the ground truth. IoU measures how close the generated image to the ``real image'' (the higher the better, maximum =1). Functional metrics are summarized in Table \ref{tab:funmetric}.
\begin{table*}[th!]
    \centering
    \caption{\textbf{Functional Metrics (Scores) Summary}}
    \label{tab:funmetric}
    \begin{tabular}{p{0.5\linewidth}p{0.2\linewidth}p{0.2\linewidth}}
    \hline
    
    Name & Abb. & Enhancement Direction \\ \hline
    Inception Score                & \textbf{IS}     & higher is better     \\ 
    Frechet Inception Distance     & \textbf{FID}    & lower is better      \\ 
    Peak Signal to Noise Ratio$^{a}$     & \textbf{PSNR}   & higher is better     \\
    Mean Pixel accuracy$^{a}$              & \textbf{mPA}  & higher is better     \\
    Intersection over Union$^{a}$        & \textbf{IoU}    & higher is better  \\
                                   
    \hline
    \multicolumn{3}{l}{\footnotesize{$^a$maximum value is 1 and requires ground truth}} \\ 
    \hline
    \end{tabular}
    \end{table*}

Performance metrics measures the efficiency of the model optimization. The efficiency of a model refers to its speed, power and used area. These metrics depends on both software model and hardware architecture and used technology. Focusing on compression techniques, our main metrics is the compression ratio, no. of macs (multiply-accumulate operations) and throughput as listed in Table \ref{tab:permetric}.
\begin{table*}[th!]
    \centering
    \caption{\textbf{Performance Metrics Summary}}
    \label{tab:permetric}
    \begin{tabular}{p{0.5\linewidth}p{0.2\linewidth}p{0.2\linewidth}}
    \hline
    Name & Abb. or Unit & Enhancement Direction \\ \hline
    Compression Ratio                     & \textbf{CR}   & higher is better  \\
    No. of multiply and accumulate ops    & \textbf{\#MACs}   & lower is better  \\ 
    Throughput$^*$                                & \textbf{output/s}    & higher is better \\
    \hline
    \end{tabular}
    \end{table*}

\subsection{Memory Compression}\label{sec:MemoryMinimization}

\textit{\textbf{What is memory compression ?} } Any DL Model needs three items to reside in the memory: 1) Model Architecture (control flow or computational graph), 2) Model Parameters (weights and biases), and 3) Inputs (activations). Compressing memory means minimizing any/all the above three components.

\textit{\textbf{What are the advantages of using memory compression  techniques ? }} It improves storage space required to store model leading to cross-layer optimization by allowing several layers to fit into the chip memory at once or even storing the whole model on embedded system. It also decreases off-chip data-transfer leading to higher throughput and lower power usage.

\begin{figure}[th!]
    \centering
    \includegraphics[width=0.9\textwidth]{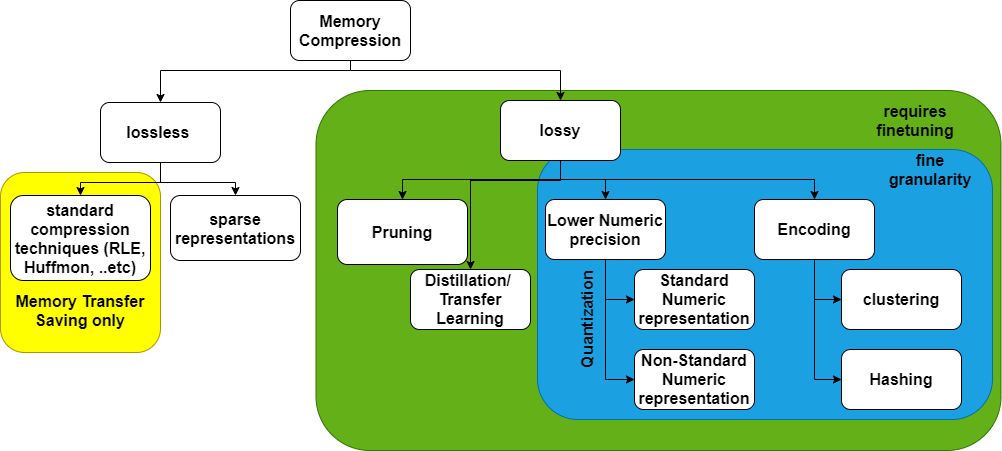}
     \if\myacm1 
        \Description{MemoryFootprintDiagram} 
     \fi
    \caption{\textbf{Classification of Memory Compression Techniques }}
    \label{fig:MemoryFootprintDiagram}
    \end{figure}

\textit{\textbf{What is the classification of memory compression techniques ?} } 
Memory Compression can be done using several techniques as shown in fig.  \ref{fig:MemoryFootprintDiagram}. Compression techniques are classified into two main categories: a) lossless compression techniques and b) lossy ones. As the name stated, lossy compression introduces some losses. That's why finetuning or retraining of the compressed model is advised with those techniques. On the other hand, lossless compression techniques require extra SW and/or HW support to revert compression or apply computation on sparse matrices. We also classify the techniques according to the granularity, a coarse granularity would consider compressing the model architecture, while a finer granularity would consider compressing data used by the model (both weights and activation). Combining several techniques together is always an option, but a careful eye needs to watch the quality of the solution. The rest of this section will focus on lossy compression techniques. It will explain each technique and show the latest work using them on GANs generators. Lossless techniques are general ones that can be used with different applications and are orthogonal to lossy techniques, thus why we choose to focus on lossy techniques.

\subsubsection{Pruning}
\begin{figure}[th!]
   \centering
   \includegraphics[width=0.9\textwidth]{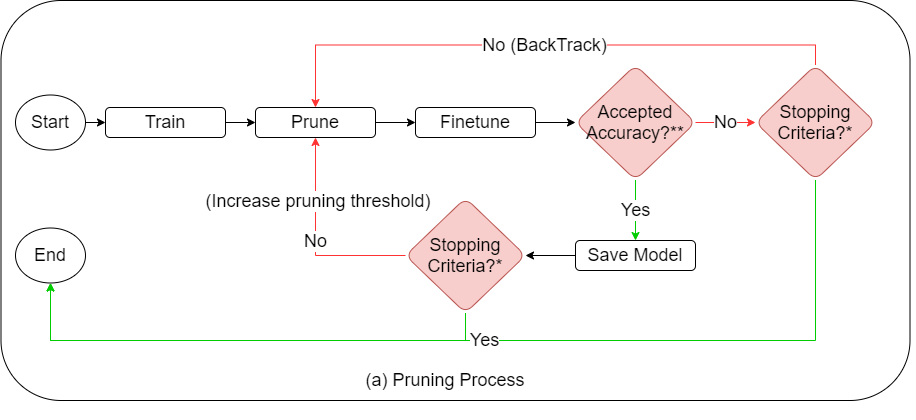}
    \if\myacm1 
       \Description{Pruning and Distillation charts} 
    \fi
   \caption{\textbf{Pruning Process Flowchart }}
   \label{fig:PruneDistillDiagram}
   \end{figure}

It is the process of eliminating parts of the network model to make it smaller without much loss of results quality. Pruning is usually done after the model is trained and then the model is finetuned to adjust the remaining weights as seen in fig.  \ref{fig:PruneDistillDiagram}. 

\begin{figure}[th!]
   \centering
   \includegraphics[width=0.9\textwidth]{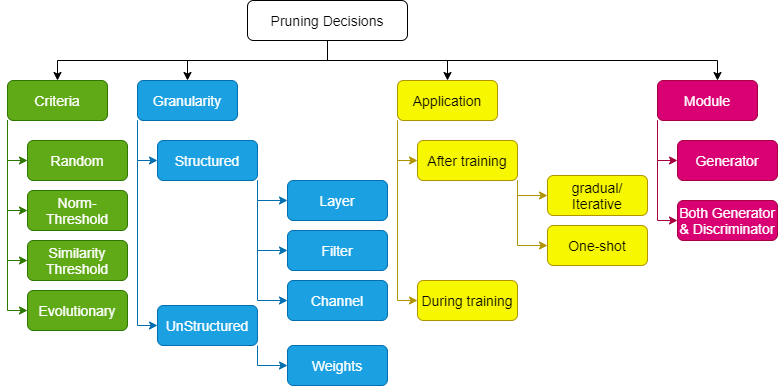}
    \if\myacm1 
       \Description{Pruning Classification} 
    \fi
   \caption{\textbf{Pruning decision: Four decisions to define any pruning technique}}
   \label{fig:PruningClassification}
   \end{figure}

Pruning is defined by 4 main decisions as seen in fig. \ref{fig:PruningClassification}. First, decision is the pruning criteria or in another words ``how to choose the part to be pruned''. Criteria could be totally random using trial and error or based on the selected element norm using some certain threshold a.k.a. norm-threshold. A similarity threshold also could be used when there are two activations (feature-maps) very similar to each other, one of them can be removed as it introduces no new information. And Finally, Evolutionary algorithms like genetic algorithm can be used to choose the elements to be pruned. 

The second decision is the pruning granularity. The pruning could be unstructured which means eliminating individual elements from weight/bias matrices. This kind of pruning leaves the network structure as it is, but it converts its weight matrices to sparse matrices which could be further compressed. On the other hand, Structured pruning eliminates components from the network leaving it slimmer (less channels or filters) or shorter (less layers). 

A third decision is the application time or ``when to apply pruning''. As mentioned previously, most work apply pruning as post-processing step, however, lately several techniques are introduced to apply pruning while training as will be explained later in this subsection. The post-processing pruning can further be classified to gradual or iterative pruning where the pruning starts with a small threshold and increases it gradually to avoid accuracy loss. On the contrary, the one-shot pruning is manually finetune network according to some criteria. \footnote{One Shot learning is usually having different meaning; it will be explored in knowledge distillation section in Network Architecture search}

A final decision is the module or ``which module should be pruned''. Pruning Generator part of the network is usually the target, however due to instability of training and hence finetuning, some works suggested to prune the discriminator as well to avoid overpowering the pruned generator while finetuning.

\begin{figure}[th!]
   \centering
   \includegraphics[width=0.9\textwidth]{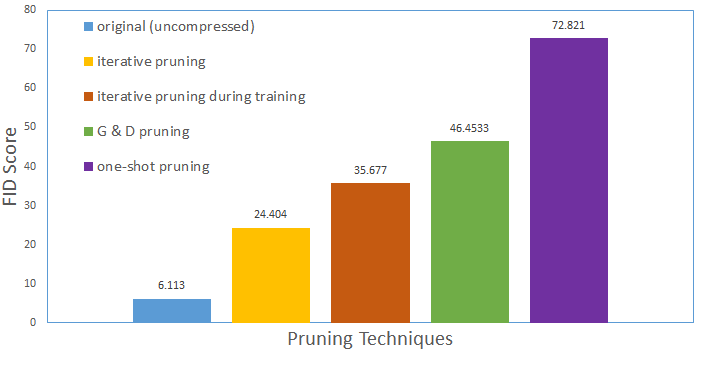}
    \if\myacm1 
       \Description{Pruning and Distillation charts} 
    \fi
   \caption{\textbf{FID scores of different pruning techniques for StarGAN, This shows the failure of pruning techniques to preserve the required quality compared to the original non-pruned version of the algorithm.}}
   \label{fig:PruningTechniques}
   \end{figure}

In a study made by Chong \& Jeff \cite{GANCompression_whyothersFail}, they showed that the quality of results degraded significantly using thresholding pruning as seen in fig.  \ref{fig:PruningTechniques}. Thresholding pruning is eliminating the element undertest if it is below a certain threshold. The element could be a single weight, a filter, or a channel. They implemented several pruning techniques on StarGAN\cite{imgtoimg_choi2018stargan}. They implemented iterative pruning after training \cite{GANCompression_whyothersFail}(e), iterative pruning during training \cite{GANCompression_whyothersFail}(f), pruning both generator and discriminator \cite{GANCompression_whyothersFail}(n) and one-shot pruning after training \cite{GANCompression_whyothersFail}(d). It worth noting that pruning during training resulted in a lower quality than pruning after training. This indicates that the model fails to converge. They also presented other non-pruning techniques which we will explore later.

Instead of using thresholding in pruning, Shu et al. proposed using evolutionary algorithm like genetic algorithm (GA) to compress the cyclic networks like CYCLEGAN \cite{prune_GAN_shu2019co}. The idea is to represent the generator as a bitstream where each bit corresponds to a filter if the bit = 0 then the filter is pruned. The  GA fitness function is a function in three criteria : a) the size of the network, b) the compression distance, and c)the cycle loss. The compression distance is the mean square error between the discriminator output for compressed and uncompressed generators. Whereas the cycle loss is a special loss in training paired images as explained in \cite{imgtoimg_zhu2017unpaired}. GA achieved a compression ratio between 3.54x $\sim$ 5.7x compression ratio on cyclegan. GA pruning achieved 0.542 mean pixel accuracy compared to 0.218 using  thresholding in pruning. It also achieved a better FID score by average of 30 points  compared to  thresholding pruning while only degrading FID than the original none-pruned with 8.5 points (calculated using 4 datasets on CYCLEGAN).

Song et al. presented overlapping pruning with training instead of the ordinary train-prune-finetune approach \cite{SPGAN_song2020sp}. In his work, the author adopted the train-expand-prune approach. He started by training a small network (called seed network), then progressively expanded it by adding more width (filters) to the network. Then similar filters are pruned and the whole network is fine tuned. They scored 1.25x less flops than baseline GAN.
   
\begin{table}[ht]
    \centering
    \caption{\textbf{Summary of GAN pruning work}}
    \label{tab:pruningComparison}
    \begin{tabular}{lcccc}
    \hline
    Work/Aspect    & Criteria     & Granularity   & Application     & Module    \\ \hline
    \cite{GANCompression_whyothersFail}(e)  & threshold    & Unstructured & After training  & Generator \\
    \cite{GANCompression_whyothersFail}(f)     & threshold    & Unstructured & during training & Generator \\
    \cite{GANCompression_whyothersFail}(n)      & threshold    & Unstructured & After training  & Generator \\
    \cite{GANCompression_whyothersFail}(d)     & threshold    & Unstructured & After training  & Both      \\
    \cite{prune_GAN_shu2019co}  & evolutionary & structured   & After training  & Generator \\
    \cite{SPGAN_song2020sp} & similarity   & structured   & during training & Generator \\\hline
    
    \end{tabular}
    \end{table}

Although many combinations of pruning techniques have not been explored in pruning as seen in the summary table.\ref{tab:pruningComparison}, the current results indicate that pruning alone is not enough and there is a significant loss in quality as shown in fig. \ref{fig:PruningTechniques}. This failure is attributed to the following  reasons:  1) The high resolution of the generator output compared to discriminator models makes it more sensitive to noise, 2) The generator evaluation metrics are more subjective than objective, and 3) the training of GANs is unstable and a careful care should be taken to avoid discriminator over-powering the generator.  To overcome those challenges, a more general approach called knowledge distillation is used where pruning is usually a part of it.

\subsubsection{Knowledge Distillation}\label{sec:KD} 
\begin{figure}[th!]
    \centering
    \includegraphics[width=0.95\textwidth]{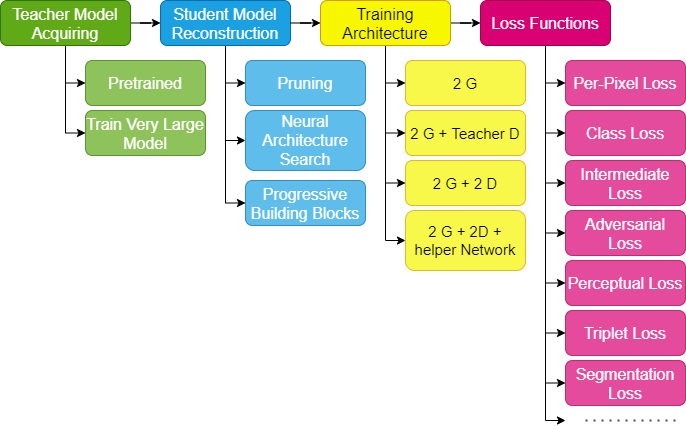}
     \if\myacm1 
        \Description{Knowledge Distillation Requirements} 
     \fi
    \caption{\textbf{Knowledge Distillation Components }}
    \label{fig:KD}
    \end{figure}

It is the transfer of knowledge acquired by the uncompressed generator (called teacher model) to a smaller model (called student model). To apply knowledge distillation, we need to define 4 main components as explained in fig.  \ref{fig:KD}. 

First component is the Teacher model. While the straight-forward approach suggests that we should have  a pretrained one that we are trying to compress, some works trained an overly large model from scratch to give them more flexibility in finding the most optimal student model.

The second component is the reconstruction of the student model. In its simplest form, pruning is used to generate the student model from the teacher model.  Student models can also be constructed using network architecture search (NAS). Or even it can be progressively constructed using sub-constructs from the teacher model.

The third component is the training architecture or building blocks. This component is concerned with which components from the teacher model will be included in the training and whether to construct a complete student GAN (both generator and discriminator) or just construct a student generator.

The last component is the loss function. The loss function determines how good and how fast the student will learn from the teacher. A lot of loss function has been introduced that will be explained below.

\begin{figure}[th!]
    \centering
    \includegraphics[width=0.95\textwidth]{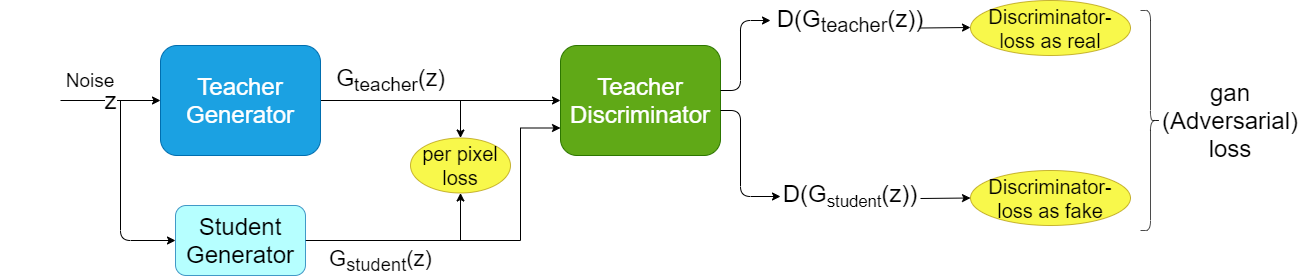}
     \if\myacm1 
        \Description{Loss functions used by Aguinaldo} 
     \fi
    \caption{\textbf{Training Architecture for work \cite{GAN_COMPRESSION_DISTILLATION} showing Per-pixel loss \& gan-loss (a.k.a. adversarial loss)}}
    \label{fig:loss2}
    \end{figure}

In \cite{GAN_COMPRESSION_DISTILLATION}, Aguinaldo et al. used knowledge distillation to train a student generator to even beat the teacher generator. They used pretrained teacher generator and discriminator as seen in fig. \ref{fig:loss2}. They devised a reconstruction loss (eq.\ref{eq:distloss}) as a joint function between  gan loss  (eq. \ref{eq:GAN_LOSS}) and  per-pixel loss (eq. \ref{eq:reconstructionloss}). First, they let the same input to both student and teacher generators. Second, they let the output of both generators  be fed into the teacher discriminator. Third, they calculate the reconstruction loss as in eq. \ref{eq:distloss} to train the compressed generator. 
\begin{align}
    \mathcal{L}_{per\_pixel} &= loss(G_{student}(z),G_{teacher}(z)) \label{eq:reconstructionloss} \\
    \mathcal{L}_{recon} &= \mathcal{L}_{gan} + \lambda \mathcal{L}_{per\_pixel} \label{eq:distloss}
\end{align}
where $z$ is the noise used as input to generators, $\mathcal{L}_{gan}$ is the adversarial gan loss from eq.\ref{eq:GAN_LOSS}, and $\lambda$ is a weighting parameter between both losses. The authors used a very large teacher to guide the small network leading 1669x compression ratio while retaining 83\% of the teacher’s Inception Score on MNIST. However, the produced images were very blurred at this compression rate. Yet what makes this work standout is that the distilled networks using this method always beat the trained-from scratch networks of the same size.

\begin{figure}[th!]
    \centering
    \includegraphics[width=0.95\textwidth]{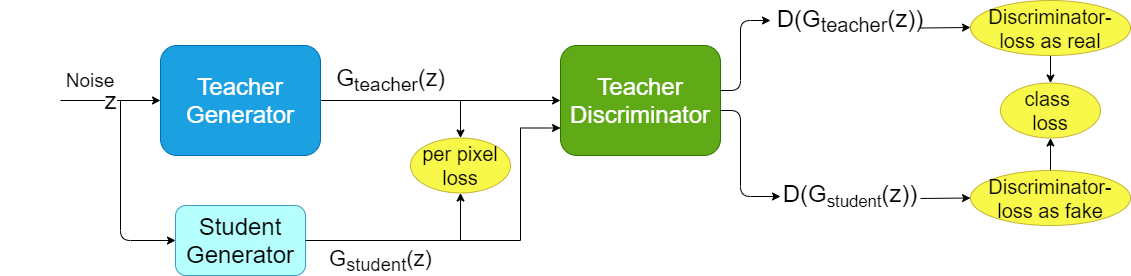}
     \if\myacm1 
        \Description{Loss functions used by Chong & Jeff} 
     \fi
    \caption{\textbf{Training Architecture for work \cite{GANCompression_whyothersFail} showing Per-pixel loss \& Class-loss}}
    \label{fig:loss1}
    \end{figure}

In \cite{GANCompression_whyothersFail}, Chong \& Jeff also used a pretrained generator and discriminator to train a student generator as shown in fig. \ref{fig:loss1}. They measured the loss as follows: first they measured the per-pixel loss as the loss between the two generators as in eq. \ref{eq:reconstructionloss}. Per-pixel loss is usually the mean square error loss, but it could be anything else. Additionally, they measured class-loss as the loss between the discriminator output of the student and teacher generators as seen in eq. \ref{eq:discreconstructionloss}. The total loss is the weighted sum of the previous two losses as shown in eq. \ref{eq:whyfail}.
\begin{align}
    \mathcal{L}_{class\_loss} &= loss(D_{teacher}(G_{student}(z)),D_{teacher}(G_{teacher}(z))) \label{eq:discreconstructionloss}\\
    \mathcal{L}_{total_1    } &= \mathcal{L}_{per\_pixel} + \lambda \mathcal{L}_{class\_loss} \label{eq:whyfail}
\end{align}

\begin{figure}[th!]
    \centering
    \includegraphics[width=0.95\textwidth]{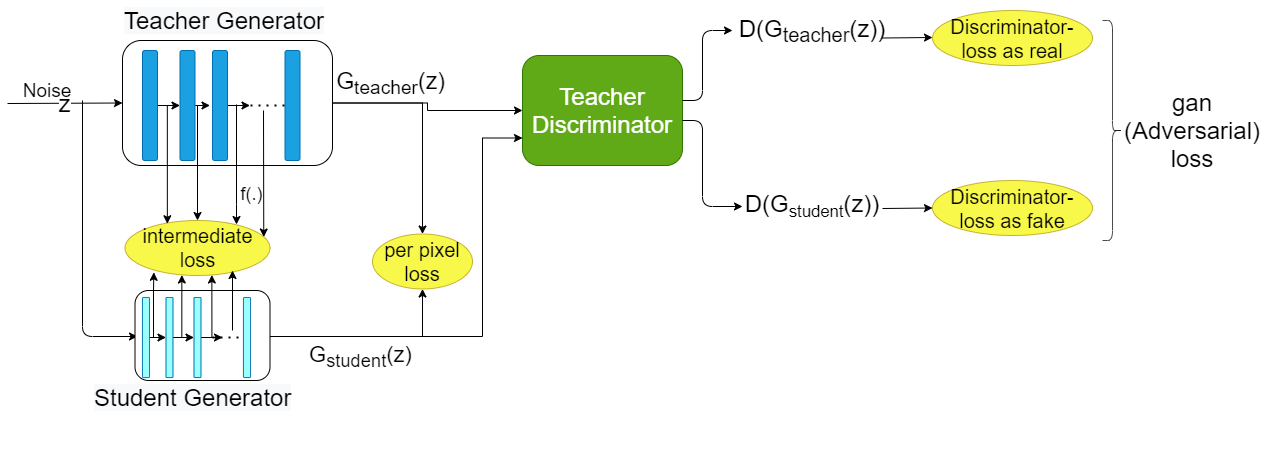}
     \if\myacm1 
        \Description{Loss functions used by Chong & Jeff} 
     \fi
    \caption{\textbf{Training Architecture for work \cite{KD_GAN_COMPRESSION_NAS,KD_GAN_fu2020autogan} showing intermediate distillation loss, Per-pixel loss \& gan-loss.  }}
    \label{fig:loss3}
    \end{figure}

Similarly, In \cite{KD_GAN_COMPRESSION_NAS}, they used $\mathcal{L}_{gan}$ function using the teacher discriminator. Additionally, they added another loss term representing intermediate distillation loss. Intermediate distillation loss is the loss between two corresponding inner-layers outputs as in eq. \ref{eq:intermediatedist}. If the data is  paired, then they calculate $\mathcal{L}_{per\_pixel}$ between student generated image and the paired image, else they use the output of the teacher generator to calculate $\mathcal{L}_{per\_pixel}$. The training architecture is shown in fig. \ref{fig:loss3} while the final loss function is seen in eqn. \ref{eq:distillnas}. To construct the student model, they used a neural architecture search (NAS).  To avoid the long running time of NAS, they used one shot learning to train a variable number of networks at once. The idea of one-for-all network training (OFA) also called one-shot learning in training is best explained in \cite{OFA_NAS}. They reached a compression ratio between 4x $\sim$ 33x on varies datasets and networks.
\begin{align}\label{eq:intermediatedist}
    \mathcal{L}_{interdist} = \sum_{t=1}^{T} ||  f_t(G_{teacher_t}(x)) - G_{student_t}(x)  ||_2 
\end{align}
where $f_t$ is the mapping function between teacher inner layer and the corresponding student layer to adjust the size. $t$ is the layer number.
\begin{align}\label{eq:distillnas}
    \mathcal{L} = \mathcal{L}_{gan} + \lambda_{per\_pixel}\mathcal{L}_{per\_pixel} + \lambda_{interdist}\mathcal{L}_{interdist}
\end{align}
where different $\lambda$ are used to weight the different loss functions.

In contrast to the previous work, the work in \cite{KD_GAN_fu2020autogan} suggested to use more constrained search for the student model instead of unconstrained NAS. They built DART, an AutoML-Like framework for GANs to perform differential search \cite{liu2018darts} for the operators at each layer and layer width . They constrained the first and last layers to be like famous models’ architecture like  Cycle-GAN in style-transfer tasks and ESRGAN\cite{wang2018esrgan} in super-resolution tasks.

\begin{figure}[th!]
    \centering
    \includegraphics[width=0.95\textwidth]{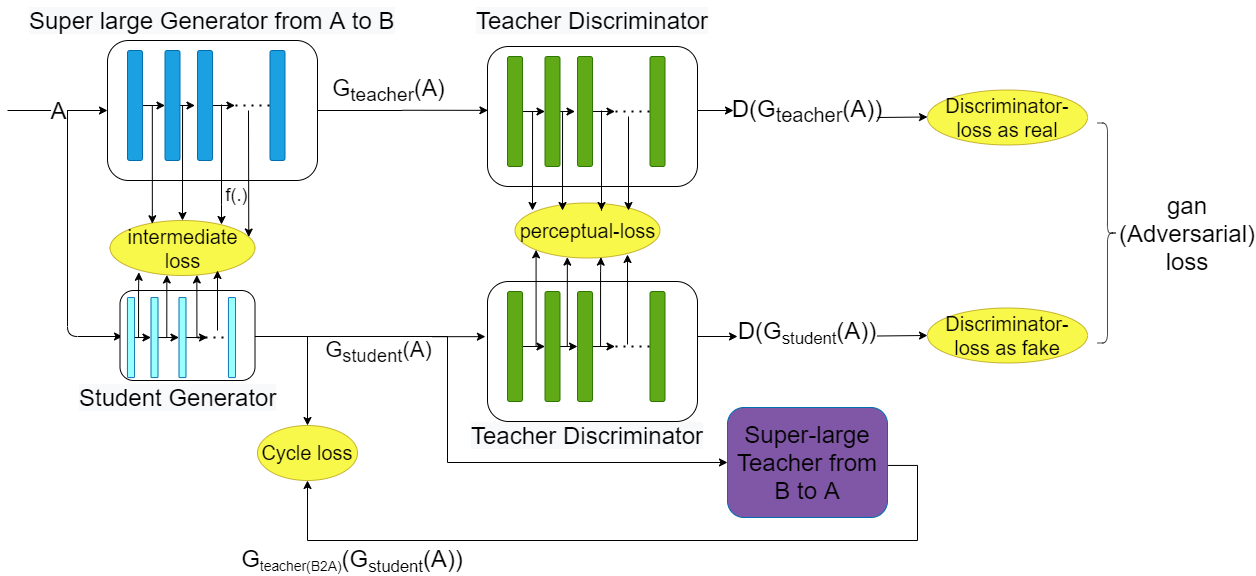}
     \if\myacm1 
        \Description{Loss functions used by Chong & Jeff} 
     \fi
    \caption{\textbf{Training Architecture for work \cite{KD_GAN_CATjin2021teachers} showing intermediate distillation loss, perceptual-loss \& gan-loss}}
    \label{fig:loss4}
    \end{figure}

In the work made by Qing et al., they reconstructed the teacher network to be a large supernet for image-to-image translation. Thus, pruning and distilling such a large network would lead to a more efficient student network\cite{KD_GAN_CATjin2021teachers}. The student network is made by pruning channels of the teacher generator. They used aggregated 4 loss-function. The training architecture is shown in fig. \ref{fig:loss4}. First, they used intermediate distillation using kernel alignment function to map the corresponding intermediate layer sizes. Second, they used perceptual loss, which is the loss between intermediate features in the discriminator between real(teacher) and student images as seen in eq. \ref{eq:perceptualloss}. Third, they used gan adversarial loss as all other GANs. Eventually, they used the cyclic-loss since it is image-to-image translation task. They cycle loss is the loss of converting from domain A to B then back to A'. The difference between A and A' is the cycle loss. This reconstruction method optimized the time required to construct student network by 10,000x compared to unconstrained NAS with better FID scores.
\begin{align}\label{eq:perceptualloss}
    \mathcal{L}_{perc} = \sum_{t=1}^{T} \frac{1}{N_t} || D_t(x) - D_t(G_{student}(z)) ||_1
\end{align}
where $x$ is real image or teacher image$G_{teacher}(z)$, and $T$ is the number of layers and $N$ is the total number of elements in each layer.

\begin{figure}[th!]
    \centering
    \includegraphics[width=0.95\textwidth]{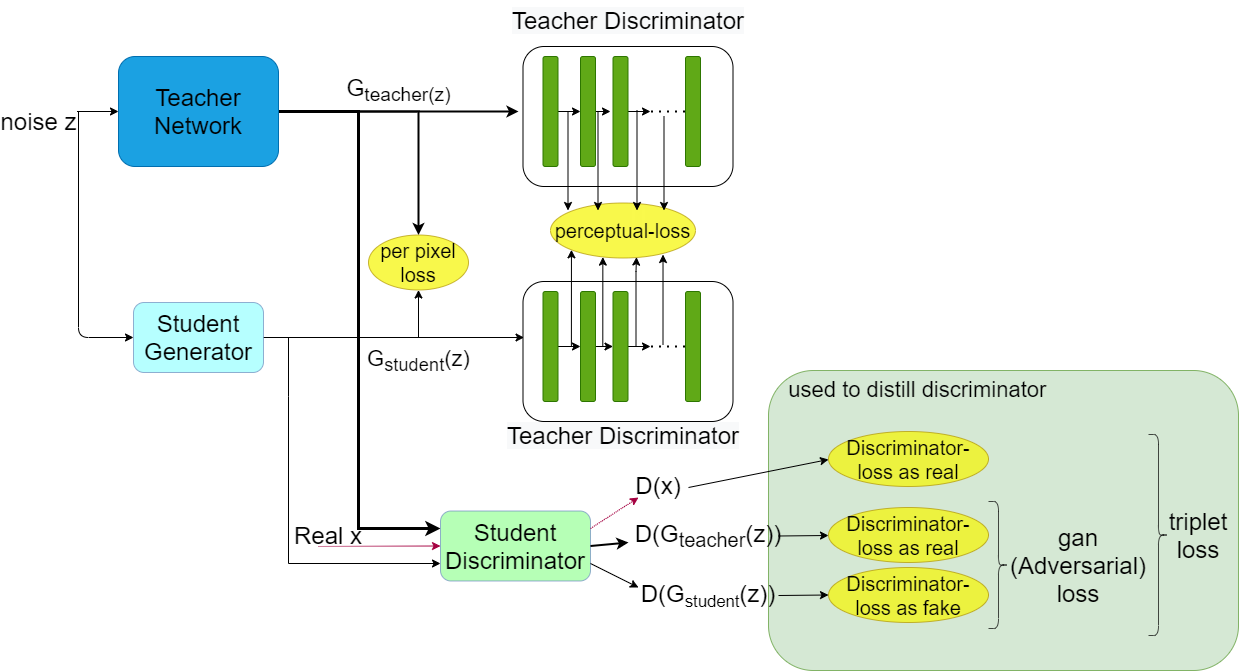}
     \if\myacm1 
        \Description{Loss functions used by Chong & Jeff} 
     \fi
    \caption{\textbf{Training Architecture for work \cite{KD_GAN_chen2020distilling} showing both generator and discriminator distillation}}
    \label{fig:loss5}
    \end{figure}

Another work by Chen et al.  used intermediate distillation but on discriminator instead of generator as shown in fig. \ref{fig:loss5}. They used two discriminators instead of one\cite{KD_GAN_chen2020distilling}, so they distill both generator and discriminator. The distillation of the generator required two loss functions: perceptual-loss, per-pixel loss and adversarial-loss. To distill the discriminator, they used the adversarial loss and  introduced another loss term, which is  triple loss. The idea of this loss term is to consider that the distance between teacher generated, and student generated images will be smaller than the distance between real images and student images, thus they added an extra parameter to weight the two differences in the loss function.

In \cite{KD_GAN_liu2021content}, the authors combined intermediate distillation loss,   perceptual loss and per-pixel loss. They also used a content aware approach to enhance distillation. This is done by detecting the areas of interest using an auxiliary segmentation network (content-aware network) and mask the corresponding images before calculating loss leading to 10x to 11.5x compression ratios compared to several original GANs with FIDs $\sim$ 7.5 for FFHQ dataset (original FID 2.7 $\sim$ 4.5). Zhang et al.  also combined intermediate distillation loss from discriminator and generator in their work presented PKDGAN\cite{KD_GAN_PKDGAN}. However, it is only applied on novelty detection, so a further work needs to compare it with other GAN methods. 
\begin{figure}[th!]
    \centering
    \includegraphics[width=0.7\textwidth]{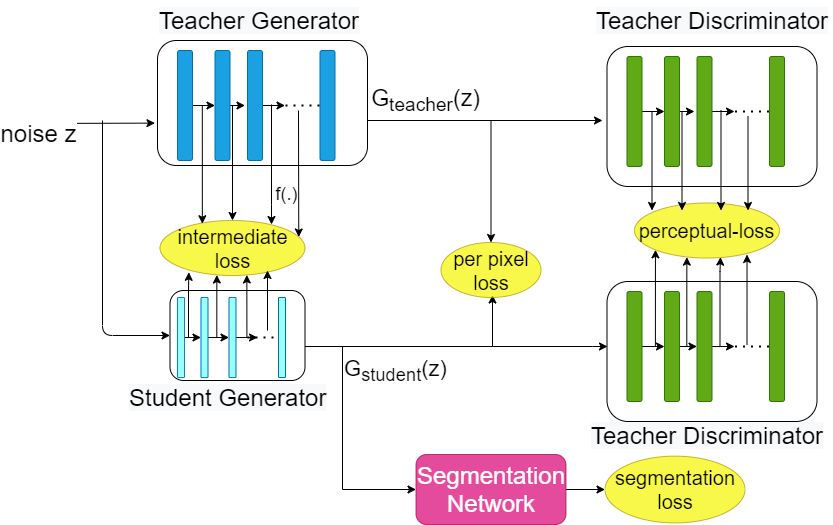}
     \if\myacm1 
        \Description{Loss functions used by Chong & Jeff} 
     \fi
    \caption{\textbf{Training Architecture for work \cite{KD_GAN_liu2021content} showing both generator and discriminator distillation and using segmentation network to allow content-aware distillation}}
    \label{fig:loss6}
    \end{figure}

In a work done by Haotoa et al., They presented a unified framework called ganslimming to stack several memory compression techniques together \cite{GAN_Slimming}. They used distillation to compress the generator. The student generator is automatically and adaptively generated from the teacher generator by channel-pruning and quantization. They used the normalization layer scale parameter to guide the pruning during training and adapted the adversarial-loss, per-pixel loss and the added normalization scale parameter as the combined loss function.

\begin{table}[ht]
    \centering
    \caption{\textbf{Summary of GAN Distillation works}}
    \label{tab:KDsummary}
    \begin{tabular}{lllll}
        \hline
    work/        & Teacher  & Student  & Training   & Loss    \\
    Component       &  Model &  Recons. &  Architecture  &  Function   \\
\hline
    \cite{GANCompression_whyothersFail}  & pretrained    & pruning  & 2G+ Teacher D & \textcolor{green}{per-pixel}       \\
& & & & \textcolor{blue}{per-class}       \\
    \hline{\cite{GAN_COMPRESSION_DISTILLATION}} & pretrained    & pruning  & 2G+ Teacher D & \textcolor{green}{per-pixel}      \\
& & & & \textcolor{red}{adversarial loss}\\
    \hline{\cite{KD_GAN_COMPRESSION_NAS}}      & pretrained    & NAS      & 2G+ Teacher D & \textcolor{green}{per-pixel}      \\
& & & & \textcolor{red}{adversarial loss}\\
& & & & \textcolor{cyan}{intermediate loss}    \\
    \hline{\cite{KD_GAN_fu2020autogan}}& pretrained    & NAS    & 2G+ Teacher D & \textcolor{green}{per-pixel}      \\
& & (DART) & & \textcolor{red}{adversarial loss}\\
& & & & \textcolor{cyan}{intermediate loss}    \\
    \hline{\cite{KD_GAN_CATjin2021teachers}}    & super-large   & pruning  & 2G+ Teacher D & \textcolor{orange}{perceptual loss} \\
& & & & \textcolor{red}{adversarial loss}\\
& & & & \textcolor{cyan}{intermediate loss}    \\
& & & & cycle-loss \cite{imgtoimg_zhu2017unpaired} \\
    \hline{\cite{KD_GAN_chen2020distilling}}    & pretrained    & pruning  & 2G+ 2D  & \textcolor{green}{per-pixel}     \\
& & & & \textcolor{orange}{perceptual loss}    \\
& & & & \textcolor{red}{adversarial loss}    \\
& & & & Triplet-loss    \\
    \hline{\cite{KD_GAN_liu2021content}}        & pretrained    & pruning  & 2G + D& \textcolor{green}{per-pixel}      \\
& & & + Segmentation  & \textcolor{orange}{perceptual loss}    \\
& & &   Network      & \textcolor{cyan}{intermediate loss}    \\
& & & & segmentation-loss      \\

   \hline\cite{GAN_Slimming}   & pretrained & pruning & 2G+ Teacher D     & \textcolor{green}{per-pixel} \\   
   & & & & \textcolor{red}{adversarial loss} \\
   & & & & normalization-loss\cite{liu2017learning} \\
    \hline 
    \end{tabular}
    \end{table}

\begin{figure}[th!]
    \centering
    \includegraphics[width=0.9\textwidth]{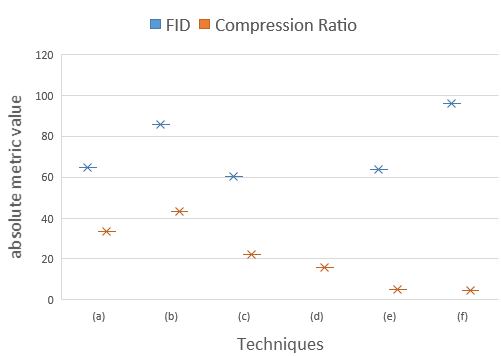}
     \if\myacm1 
        \Description{Knowledge Distillation FID scores} 
     \fi
    \caption{\textbf{Comparison between different distillation and pruning techniques using cyclegan model and horse2zebra dataset. (a),(b),(c),(d),(e),(f) represents the works in \cite{KD_GAN_COMPRESSION_NAS},
    \cite{KD_GAN_fu2020autogan},
    \cite{KD_GAN_CATjin2021teachers},
    \cite{KD_GAN_chen2020distilling},
    \cite{GAN_Slimming},
    \cite{prune_GAN_shu2019co} respectively. (f) \cite{prune_GAN_shu2019co} is a pruning technique, while all the others are distillation. In (e), we only reported the distillation results and omitted the quantization effect for a fair comparison.  Work (a),(c) had stronger teacher than the rest. In (c), we estimated the compression ratio as the ratio between the number mac operations.}}
    \label{fig:KDcomp}
    \end{figure}

Table \ref{tab:KDsummary} summarizes the above-mentioned works and how their four main components are selected. While fig. \ref{fig:KDcomp} shows the difference performance of the above techniques with respect to each other and with a sample from pruning techniques as well. For a fair comparison, we used only techniques with Cyclegan and horse2zebra dataset. The axes names in the fig.  a, b, c, d, e, f represent the following works \cite{KD_GAN_COMPRESSION_NAS},
\cite{KD_GAN_fu2020autogan},
\cite{KD_GAN_CATjin2021teachers},
\cite{KD_GAN_chen2020distilling},
\cite{GAN_Slimming},
\cite{prune_GAN_shu2019co} respectively. 
\cite{KD_GAN_chen2020distilling} didn\'t report the FID score in his paper, that\'s why it is omitted from the graph. In works \cite{KD_GAN_COMPRESSION_NAS,KD_GAN_CATjin2021teachers}, their teacher model FID score was better than the others. 
This justifies why they have much better FID score because they had a better teacher. However, Wang et al. in \cite{GAN_Slimming} managed to score very close to them despite starting from a weaker teacher although its compression ratio is one of the lowest compared to others. It worth noting that the pruning technique done by \cite{prune_GAN_shu2019co}  has the worst compression ratio and the worst FID score, which is consistent with our conclusion in pruning section, that pruning alone is not enough.

\subsubsection{Lowering numeric precision} \label{sec:LowerPrecision}
\begin{figure}[th!]
    \centering
    \includegraphics[width=0.8\textwidth]{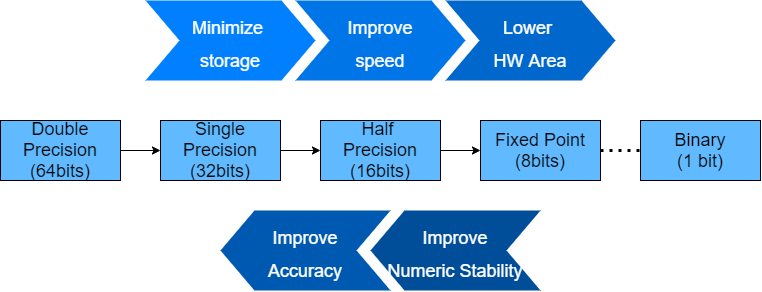}
     \if\myacm1 
        \Description{number representation impact} 
     \fi
    \caption{\textbf{Impact of precision on performance metrics }}
    \label{fig:numberrepresentationimpact}
    \end{figure}

As the name stated, it is using less bits to represent a number. Lowering numeric precision is not unique to Machine Learning. As the precision increases the accuracy and numerical stability increases. On the other hand, as the precision decreases, the speed, memory footprint and hardware area get improved \ref{fig:numberrepresentationimpact}. Despite that, the relation is not linear, and it differs according to the application. 

\begin{figure}[th!]
    \centering
    \includegraphics[width=0.95\textwidth]{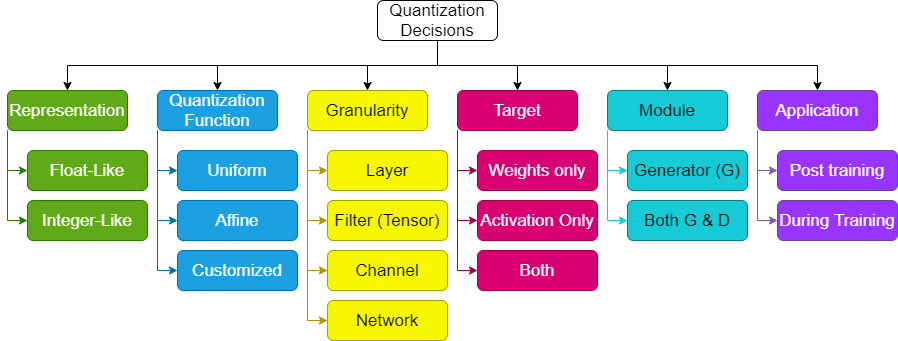}
     \if\myacm1 
        \Description{quantization decisions} 
     \fi
    \caption{\textbf{Quantization Decisions}}
    \label{fig:quantizationComponents}
    \end{figure}

As mentioned earlier, GAN generators are more sensitive to precision due to the resolution of the output. Thus, in this section, we will show the impact of such optimization on generators and explore different numeric formats. Changing numeric precision can be done by using standard formats like single precision (float32), half precision (float16) or fixed-point representation which can take many forms depending on the place of the fixed-point. Those standard format has more adoption in hardware since they are "standard". On the contrary, Non-standard formats use out-of-the box ideas like Bfloat\footnote{Although Bfloat is not IEEE standardized, however google TPU support for it makes it common to use and supported by different optimized deep learning libraries.}, FlexPoint, \dots etc. Those out-of-the box format needs dedicated hardware support to be used.
Lowering numeric precision to Fixed-point integer is usually called quantization in literature whenever the quantization follows the affine mapping as in eq. \ref{eq:affine}. The advantage of using affine transformation is that the multiplication and addition can be carried out without the need to revert the mapping \cite{gogo_quantization}. Other types of quantization requiring reverting back the conversion before calculations  is considered as a type of encoding.
\begin{align}\label{eq:affine}
    quantized\_number = \frac{full\_precision - zero\_point}{scale\_factor}
\end{align}
To perform quantization, we need to define the necessary decisions shown in fig. \ref{fig:quantizationComponents}. First decision is the general numeric format. Whether the number format is float-like, or integer-like. Float-like format is  (sign,exponent,mantissa) format. The integer-like format like fixed point format requires (Integer, Scale factor, Zero-point) format.
Second decision is the quantization function. Determining the quantization function determines how the full-precision number is converted to the quantized one as shown in eq.\ref{eq:affine}. The quantization could be uniform or could give a higher priority to certain range depending on the data distribution using log or tanh function. 

Third to Fifth decisions are about what part of the network to apply quantization to.
Starting by the granularity whether the same quantization should be applied to all layers, or each layer should have different quantization. The quantization could be per-tensor, or per-channel or per-layer or per network (G has different quantization than D b). Fourth decision is whether to quantize weights or activations or both. If only one type of data is quantized, then the saving will be in storage only not in calculation as it must cast back to the largest of the two formats. And the fifth one is whether to quantize both networks or the generator only.

Last decision is when to apply the quantization. Post-training quantization requires finetuning while during training or called ``Training-aware quantization'' doesn\'t need any additional finetuning. 

Wang introduced a method to quantize GAN called QGAN\cite{QGAN}. In his study, he showed that normal quantization methods like (uniform, log, tanh) are not sufficient for a stable and convergent GAN in a very low-bit quantization. Moreover, $G$ and $D$ sensitivity to quantization is different. Though eventually, only the generator is needed, but finetuning it needs a balanced discriminator. Thus, quantizing both discriminator and generator leads to a convergent finetuning. This leads to proposing multi-precision quantization for $G$ and $D$ separately from each other. Then he used the EM algorithm to find the optimal  $zero\_point and scale\_factor$ parameters from eq. \ref{eq:affine}. Quantized weight is calculated from the eq.\ref{eq:wq}. The EM algorithm tries to minimize the mean square error between the non-quantized weight and the quantized one.
\begin{align}
    W_q = {scale\_factor}*round(quantized\_number)+zero\_point \label{eq:wq}
\end{align}
where $scale\_factor, quantized\_number, zero\_point$ are same parameters from eq.\ref{eq:affine}. $W_q$ is the quantized weight, $round(quantized\_number)$ is the integer part of the fixed-point number that will be used in the calculation. 

QGAN has been applied to several gans like DCGAN\cite{DCGAN}, WGAN\cite{WGAN-GP} and LSGAN \cite{LSGAN}. With a quantization to 1-4 bits, QGAN achieved a compression ratio from 8x up to 32x with a small loss in inception score.

A study made by Deng et al. using a PATCH-GAN like generator to reconstruct face images showed that as the number of bits decreases,  the Peak Signal-to-Noise Ratio (PSNR) gets worse while the memory footprint improves, which is not surprising\cite{QMGAN}. In their study for Quantized GAN for Mobiles (QMGAN), They found that 32-bit representation (single precision) has $\sim$ 2.5x improvement in PSNR over the 1-bit quantized (binary) network. However, the 1-bit quantized network has better memory size by 35x over single precision. They tried different values for quantization, what worth noting is that the PSNR of 32-bit is almost the same as the PSNR of 6-bits while the 6-bit has around 5.4x memory size improvement. 

Haotao et al. continued their work on ganslimming\cite{GAN_Slimming} by applying both quantization and knowledge distillation. He used the uniform quantization on both activations and weights on the generator model. They unified the quantization on all layers so that it would be HW friendly. They performed 4x $\sim$8x compression ratios on style-transfer problems with a very competitive result.

In \cite{ApGAN_COM_PIM_DF_MEM}, the authors of ApGAN used memory compression on a ReRam accelerator \footnote{ReRam accelerator  is processing in memory using analog crossbar circuit}. They quantized all weights and activations to 1-bit (sign-bit)which simplified MACs to just ANDing followed by ORing. This technique decreases the weights storage size. However, that comes at a cost of accuracy and speed of convergence. An experiment done by \cite{ApGAN_COM_PIM_DF_MEM} compared fully-binarized DCGAN to full-precision, after 20 epochs the loss of binarized version is 3x the loss of full precision. For that reason, instead of binarizing all layers, they used variable layers quantization based on the data redundancy measure. Data redundancy measure  is defined as $c_i- h_i*w_i$ where $i$ is the layer number $c_i$ is the number of channels of layer input $i$, and $h_i,w_i$ are the input height and width respectively of the $i^th$ layer. A negative redundancy measure indicates a high sensitivity for quantization error; thus, it is not recommended to quantize such layers. Other layers with high redundancy measure are binarized by taking the sign bit of the weight and the average weight is considered the scale factor. Multiplication computations turns into just sign manipulation operation followed by scaling. 

In \cite{PIMTGAN}, Rakin et al. proposed TGAN, a GAN that ternarize weights to \{-1,0,1\}. The ternarization depends on the sign of the weight like ApGAN. They applied the quantization schema on both generator and discriminator in the forward path and used the backward path to update the scale factor. They achieved on average $\sim85\%$ of IS of the full-precision network.

A work by K{\"o}ster et al. proposed using non-standard numeric format called Flex-point\cite{FlexPoint}. The new format makes one shared exponent for each tensor; thus, the tensor operations are handled as if they were fixed-point operations, and an extra circuit is needed to manage exponent which is faster than the floating-point operation with different exponent for each number. Flex-point format stores the tensor as 16-bit mantissa for each element and one shared 5-bits exponent for the whole tensor.  In contrast to floating point, the exponent is shared across tensor elements, and different from fixed point, the exponent is updated automatically every time a tensor is written. By implementing several networks using, the flex-point format, the FID score of those networks was comparable to the same networks implemented using float32 and better than the ones implemented with fixed-point or float16. Flex-point has the advantage of supporting training and applying all techniques of floating-point with even faster calculation given their newly  Flex-point format. 
\begin{table}[ht]
    \centering
    \caption{\textbf{Summary of Quantization works}}
    \label{tab:Quantsummary}
    \begin{tabular}{lllllll}
    \hline
    decision          & rep. & func.   & granularity & Target   & Module & App.     \\
    work & & & & & & \\
    { QGAN \cite{QGAN}}                      & Int   & EM         & Network     & W                & G \& D   & Post   \\
    { QMGAN\cite{QMGAN}}                     & Int   & Uni    & Network     & W                & G      & Post   \\
    { ApGAN\cite{ApGAN_COM_PIM_DF_MEM}$^{+}$} & Int   & Uni    & Layer       & W                & G \& D & During \\
    { TGAN\cite{PIMTGAN}$^{+}$} & Int   & Uni    & Layer       & W                & G \& D & During \\
    { Flexpoint\cite{FlexPoint}$^+$}                 & Float     & Cust. & Tensor      & A \& W  & G \& D & During \\
    { GANslim\cite{GAN_Slimming}}             & Int   & Uni    & Network     & A \& W & G      & Post   \\

    \hline
    \multicolumn{2}{l}{\footnotesize{Int $\leftarrow$ Integer-like}}  & \multicolumn{2}{l}{\footnotesize{Uni $\leftarrow$ Uniform}} &  \multicolumn{2}{l}{\footnotesize{A $\leftarrow$ Activation}}\\
    \multicolumn{2}{l}{\footnotesize{$^+$ customized accelerator}}  & \multicolumn{2}{l}{\footnotesize{Cust $\leftarrow$ Customized}} &  \multicolumn{2}{l}{\footnotesize{W $\leftarrow$ Weights }}\\
    \hline              
    \end{tabular}
    \end{table}
	
\begin{figure}[th!]
    \centering
    \includegraphics[width=0.8\textwidth]{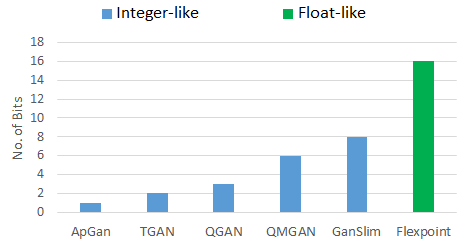}
     \if\myacm1 
        \Description{quantizedBits} 
     \fi
    \caption{\textbf{No. of quantized bits that resulted in the best performance for the GAN undertest, Each technique is tested on different GAN. The works QGAN\cite{QGAN}, QMGAN\cite{QMGAN}, GanSlim\cite{GAN_Slimming}, Flexpoint\cite{FlexPoint} and TGAN\cite{PIMTGAN}	are fully quantized. While  ApGan \cite{ApGAN_COM_PIM_DF_MEM} has mixed quantized and non-quantized layers. Also, FlexPoint, ApGAN and TGAN quantization is used during training }}
    \label{fig:quantizedBits}
    \end{figure}

Table \ref{tab:Quantsummary} summarizes the design decision taken by the different works, and because each work uses a different network it is hard to make a fair comparison using metrics like IS or FID. Fig. \ref{fig:quantizedBits} showed the no. of bits that quantization methods reported for best results.

\subsubsection{Encoding} 

It is a form of lossy compression used to minimize the data transfer using fewer bits. The data transferred to the chip is not the real data, but an index or key used to get the real data from a preloaded codebook or using a predefined hash function.

 This technique has been extensively used in deep learning like in \cite{CNN_compression_Clustering,CNN_compression_resharing,CNN_compression_hashing,CNN_compression_hashingBlock,CNN_compression_structuredhashing} . However, to the best of our knowledge, we found no work applied to GANs and it is one of the opportunities to seek.

\section{Related Work}\label{related}

Several survey papers exist about deep learning compression like \cite{RELATED_NOGAN_cheng2017survey,RELATED_NOGAN_cheng2018model,RELATED_NOGAN_Cheng2018RecentAI,RELATED_NOGAN_choudhary2020comprehensive}.
However, those papers discuss general deep learning algorithm not targeting issues specific to GANs. While some techniques mentioned in previous surveys might apply to GANs, GANs propose more challenges and opportunities. First, generative networks are sensitive to number-representation. In another words, Not all quantization techniques used for general DL models would be efficient for use with GANs. Second, GAN training suffers from instability, which makes normal compression and pruning techniques inefficient. Finally, GAN output resolution is large and correlated as opposed to normal classification or regression problems whose output are very small.

Another related work is the work done by Gou et al \cite{RELATED_NOGAN_distill_usedGAN_gou2021knowledge}. Although this work ``uses'' GAN to perform distillation, it does not consider GAN themselves for compression. While the work by Wang and Yoon \cite{RELATED_distill_usedGAN_Wang2021KnowledgeDA} mentioned briefly impact of distillation on image translation tasks using GANs, its focus was using GAN to perform distillation to other models.

\section{Conclusion \& Future work} \label{summary}

In this paper, we surveyed the lossy compression techniques used to optimize GANs. as mentioned earlier, GAN differ from other DL models because of the instability of the training and the resolution of the output.

No doubt that minimizing the memory footprint enhances storage, speed, and power efficiency of running GANs. But there are some areas like ``encoding'' that is not explored at all in GAN-domain. As seen from tables  tab.  \ref{tab:pruningComparison}, \ref{tab:KDsummary}, \ref{tab:Quantsummary} a lot of combinations can still be explored like using similarity-pruning with knowledge distillation. Also combining  losses introduced in knowledge distillation with quantization needs more exploration. 

A challenge  exists with all optimization techniques mentioned is how to unify measures (both qualitative and quantitative) and benchmark between several methods. With no unified measures, benchmarks, and platforms, we can hardly evaluate techniques compared to each other. Another opportunity exists for filling the gaps and mixing between different optimization techniques.

A future work of this survey is 1) unifying the design metrics between different designs and provide an evaluation study using several dataset, 2) explore computational optimization, 3) explore dataflow optimization, 4) study systems with combinations of optimization techniques, 5) study the impact of optimization on different platform (FPGA, ASIC, RERAMs, GPUS, ...etc.), while compression is very plausible some techniques are not hardware friendly though they give high compression ratio with low accuracy loss.  An opportunity exists in enhancing indexing methods in accelerators or building a cache like system to support clustering or hashing -based quantization. Also, the support for processing compressed elements or quantized element should be considered without uncompressing them.



\begin{backmatter}


\bibliographystyle{bmc-mathphys} 
\bibliography{main_references}      


\begin{thebibliography}{49}
\ifx \bisbn   \undefined \def \bisbn  #1{ISBN #1}\fi
\ifx \binits  \undefined \def \binits#1{#1}\fi
\ifx \bauthor  \undefined \def \bauthor#1{#1}\fi
\ifx \batitle  \undefined \def \batitle#1{#1}\fi
\ifx \bjtitle  \undefined \def \bjtitle#1{#1}\fi
\ifx \bvolume  \undefined \def \bvolume#1{\textbf{#1}}\fi
\ifx \byear  \undefined \def \byear#1{#1}\fi
\ifx \bissue  \undefined \def \bissue#1{#1}\fi
\ifx \bfpage  \undefined \def \bfpage#1{#1}\fi
\ifx \blpage  \undefined \def \blpage #1{#1}\fi
\ifx \burl  \undefined \def \burl#1{\textsf{#1}}\fi
\ifx \doiurl  \undefined \def \doiurl#1{\textsf{#1}}\fi
\ifx \betal  \undefined \def \betal{\textit{et al.}}\fi
\ifx \binstitute  \undefined \def \binstitute#1{#1}\fi
\ifx \binstitutionaled  \undefined \def \binstitutionaled#1{#1}\fi
\ifx \bctitle  \undefined \def \bctitle#1{#1}\fi
\ifx \beditor  \undefined \def \beditor#1{#1}\fi
\ifx \bpublisher  \undefined \def \bpublisher#1{#1}\fi
\ifx \bbtitle  \undefined \def \bbtitle#1{#1}\fi
\ifx \bedition  \undefined \def \bedition#1{#1}\fi
\ifx \bseriesno  \undefined \def \bseriesno#1{#1}\fi
\ifx \blocation  \undefined \def \blocation#1{#1}\fi
\ifx \bsertitle  \undefined \def \bsertitle#1{#1}\fi
\ifx \bsnm \undefined \def \bsnm#1{#1}\fi
\ifx \bsuffix \undefined \def \bsuffix#1{#1}\fi
\ifx \bparticle \undefined \def \bparticle#1{#1}\fi
\ifx \barticle \undefined \def \barticle#1{#1}\fi
\ifx \bconfdate \undefined \def \bconfdate #1{#1}\fi
\ifx \botherref \undefined \def \botherref #1{#1}\fi
\ifx \url \undefined \def \url#1{\textsf{#1}}\fi
\ifx \bchapter \undefined \def \bchapter#1{#1}\fi
\ifx \bbook \undefined \def \bbook#1{#1}\fi
\ifx \bcomment \undefined \def \bcomment#1{#1}\fi
\ifx \oauthor \undefined \def \oauthor#1{#1}\fi
\ifx \citeauthoryear \undefined \def \citeauthoryear#1{#1}\fi
\ifx \endbibitem  \undefined \def \endbibitem {}\fi
\ifx \bconflocation  \undefined \def \bconflocation#1{#1}\fi
\ifx \arxivurl  \undefined \def \arxivurl#1{\textsf{#1}}\fi
\csname PreBibitemsHook\endcsname

\bibitem{goodfellow2014generative}
\begin{bchapter}
\bauthor{\bsnm{Goodfellow}, \binits{I.}},
\bauthor{\bsnm{Pouget-Abadie}, \binits{J.}},
\bauthor{\bsnm{Mirza}, \binits{M.}},
\bauthor{\bsnm{Xu}, \binits{B.}},
\bauthor{\bsnm{Warde-Farley}, \binits{D.}},
\bauthor{\bsnm{Ozair}, \binits{S.}},
\bauthor{\bsnm{Courville}, \binits{A.}},
\bauthor{\bsnm{Bengio}, \binits{Y.}}:
\bctitle{Generative adversarial nets}.
In: \bbtitle{Advances in Neural Information Processing Systems},
pp. \bfpage{2672}--\blpage{2680}
(\byear{2014})
\end{bchapter}
\endbibitem

\bibitem{speech_bollepalli2019generative}
\begin{botherref}
\oauthor{\bsnm{Bollepalli}, \binits{B.}},
\oauthor{\bsnm{Juvela}, \binits{L.}},
\oauthor{\bsnm{Alku}, \binits{P.}}:
Generative adversarial network-based glottal waveform model for statistical
  parametric speech synthesis.
arXiv preprint arXiv:1903.05955
(2019)
\end{botherref}
\endbibitem

\bibitem{txttoimg_zhang2017stackgan}
\begin{bchapter}
\bauthor{\bsnm{Zhang}, \binits{H.}},
\bauthor{\bsnm{Xu}, \binits{T.}},
\bauthor{\bsnm{Li}, \binits{H.}},
\bauthor{\bsnm{Zhang}, \binits{S.}},
\bauthor{\bsnm{Wang}, \binits{X.}},
\bauthor{\bsnm{Huang}, \binits{X.}},
\bauthor{\bsnm{Metaxas}, \binits{D.N.}}:
\bctitle{Stackgan: Text to photo-realistic image synthesis with stacked
  generative adversarial networks}.
In: \bbtitle{Proceedings of the IEEE International Conference on Computer
  Vision},
pp. \bfpage{5907}--\blpage{5915}
(\byear{2017})
\end{bchapter}
\endbibitem

\bibitem{txttoimg_zhang2018photographic}
\begin{bchapter}
\bauthor{\bsnm{Zhang}, \binits{Z.}},
\bauthor{\bsnm{Xie}, \binits{Y.}},
\bauthor{\bsnm{Yang}, \binits{L.}}:
\bctitle{Photographic text-to-image synthesis with a hierarchically-nested
  adversarial network}.
In: \bbtitle{Proceedings of the IEEE Conference on Computer Vision and Pattern
  Recognition},
pp. \bfpage{6199}--\blpage{6208}
(\byear{2018})
\end{bchapter}
\endbibitem

\bibitem{imgtoimg_choi2018stargan}
\begin{bchapter}
\bauthor{\bsnm{Choi}, \binits{Y.}},
\bauthor{\bsnm{Choi}, \binits{M.}},
\bauthor{\bsnm{Kim}, \binits{M.}},
\bauthor{\bsnm{Ha}, \binits{J.-W.}},
\bauthor{\bsnm{Kim}, \binits{S.}},
\bauthor{\bsnm{Choo}, \binits{J.}}:
\bctitle{Stargan: Unified generative adversarial networks for multi-domain
  image-to-image translation}.
In: \bbtitle{Proceedings of the IEEE Conference on Computer Vision and Pattern
  Recognition},
pp. \bfpage{8789}--\blpage{8797}
(\byear{2018})
\end{bchapter}
\endbibitem

\bibitem{highres_wang2018high}
\begin{bchapter}
\bauthor{\bsnm{Wang}, \binits{T.-C.}},
\bauthor{\bsnm{Liu}, \binits{M.-Y.}},
\bauthor{\bsnm{Zhu}, \binits{J.-Y.}},
\bauthor{\bsnm{Tao}, \binits{A.}},
\bauthor{\bsnm{Kautz}, \binits{J.}},
\bauthor{\bsnm{Catanzaro}, \binits{B.}}:
\bctitle{High-resolution image synthesis and semantic manipulation with
  conditional gans}.
In: \bbtitle{Proceedings of the IEEE Conference on Computer Vision and Pattern
  Recognition},
pp. \bfpage{8798}--\blpage{8807}
(\byear{2018})
\end{bchapter}
\endbibitem

\bibitem{music_gansynth}
\begin{botherref}
\oauthor{\bsnm{Engel}, \binits{J.}},
\oauthor{\bsnm{Agrawal}, \binits{K.K.}},
\oauthor{\bsnm{Chen}, \binits{S.}},
\oauthor{\bsnm{Gulrajani}, \binits{I.}},
\oauthor{\bsnm{Donahue}, \binits{C.}},
\oauthor{\bsnm{Roberts}, \binits{A.}}:
Gansynth: Adversarial neural audio synthesis.
arXiv preprint arXiv:1902.08710
(2019)
\end{botherref}
\endbibitem

\bibitem{video_gan_clark2019efficient}
\begin{botherref}
\oauthor{\bsnm{Clark}, \binits{A.}},
\oauthor{\bsnm{Donahue}, \binits{J.}},
\oauthor{\bsnm{Simonyan}, \binits{K.}}:
Efficient video generation on complex datasets.
arXiv preprint arXiv:1907.06571
(2019)
\end{botherref}
\endbibitem

\bibitem{app_GAN_tryon_Fwgan}
\begin{bchapter}
\bauthor{\bsnm{Dong}, \binits{H.}},
\bauthor{\bsnm{Liang}, \binits{X.}},
\bauthor{\bsnm{Shen}, \binits{X.}},
\bauthor{\bsnm{Wu}, \binits{B.}},
\bauthor{\bsnm{Chen}, \binits{B.-C.}},
\bauthor{\bsnm{Yin}, \binits{J.}}:
\bctitle{Fw-gan: Flow-navigated warping gan for video virtual try-on}.
In: \bbtitle{Proceedings of the IEEE International Conference on Computer
  Vision},
pp. \bfpage{1161}--\blpage{1170}
(\byear{2019})
\end{bchapter}
\endbibitem

\bibitem{app_GAN_superresolution}
\begin{bchapter}
\bauthor{\bsnm{Galteri}, \binits{L.}},
\bauthor{\bsnm{Seidenari}, \binits{L.}},
\bauthor{\bsnm{Bertini}, \binits{M.}},
\bauthor{\bsnm{Uricchio}, \binits{T.}},
\bauthor{\bsnm{Del~Bimbo}, \binits{A.}}:
\bctitle{Fast video quality enhancement using gans}.
In: \bbtitle{Proceedings of the 27th ACM International Conference on
  Multimedia},
pp. \bfpage{1065}--\blpage{1067}
(\byear{2019})
\end{bchapter}
\endbibitem

\bibitem{DCGAN}
\begin{barticle}
\bauthor{\bsnm{Gao}, \binits{F.}},
\bauthor{\bsnm{Yang}, \binits{Y.}},
\bauthor{\bsnm{Wang}, \binits{J.}},
\bauthor{\bsnm{Sun}, \binits{J.}},
\bauthor{\bsnm{Yang}, \binits{E.}},
\bauthor{\bsnm{Zhou}, \binits{H.}}:
\batitle{A deep convolutional generative adversarial networks (dcgans)-based
  semi-supervised method for object recognition in synthetic aperture radar
  (sar) images}.
\bjtitle{Remote Sensing}
\bvolume{10}(\bissue{6}),
\bfpage{846}
(\byear{2018})
\end{barticle}
\endbibitem

\bibitem{imgtoimg_isola2017image}
\begin{bchapter}
\bauthor{\bsnm{Isola}, \binits{P.}},
\bauthor{\bsnm{Zhu}, \binits{J.-Y.}},
\bauthor{\bsnm{Zhou}, \binits{T.}},
\bauthor{\bsnm{Efros}, \binits{A.A.}}:
\bctitle{Image-to-image translation with conditional adversarial networks}.
In: \bbtitle{Proceedings of the IEEE Conference on Computer Vision and Pattern
  Recognition},
pp. \bfpage{1125}--\blpage{1134}
(\byear{2017})
\end{bchapter}
\endbibitem

\bibitem{InceptionScore}
\begin{bchapter}
\bauthor{\bsnm{Salimans}, \binits{T.}},
\bauthor{\bsnm{Goodfellow}, \binits{I.}},
\bauthor{\bsnm{Zaremba}, \binits{W.}},
\bauthor{\bsnm{Cheung}, \binits{V.}},
\bauthor{\bsnm{Radford}, \binits{A.}},
\bauthor{\bsnm{Chen}, \binits{X.}}:
\bctitle{Improved techniques for training gans}.
In: \bbtitle{Advances in Neural Information Processing Systems},
pp. \bfpage{2234}--\blpage{2242}
(\byear{2016})
\end{bchapter}
\endbibitem

\bibitem{FID}
\begin{bchapter}
\bauthor{\bsnm{Heusel}, \binits{M.}},
\bauthor{\bsnm{Ramsauer}, \binits{H.}},
\bauthor{\bsnm{Unterthiner}, \binits{T.}},
\bauthor{\bsnm{Nessler}, \binits{B.}},
\bauthor{\bsnm{Hochreiter}, \binits{S.}}:
\bctitle{Gans trained by a two time-scale update rule converge to a local nash
  equilibrium}.
In: \bbtitle{Advances in Neural Information Processing Systems},
pp. \bfpage{6626}--\blpage{6637}
(\byear{2017})
\end{bchapter}
\endbibitem

\bibitem{GANCompression_whyothersFail}
\begin{botherref}
\oauthor{\bsnm{Yu}, \binits{C.}},
\oauthor{\bsnm{Pool}, \binits{J.}}:
Self-supervised gan compression.
arXiv:2007.01491v2
(2020)
\end{botherref}
\endbibitem

\bibitem{prune_GAN_shu2019co}
\begin{bchapter}
\bauthor{\bsnm{Shu}, \binits{H.}},
\bauthor{\bsnm{Wang}, \binits{Y.}},
\bauthor{\bsnm{Jia}, \binits{X.}},
\bauthor{\bsnm{Han}, \binits{K.}},
\bauthor{\bsnm{Chen}, \binits{H.}},
\bauthor{\bsnm{Xu}, \binits{C.}},
\bauthor{\bsnm{Tian}, \binits{Q.}},
\bauthor{\bsnm{Xu}, \binits{C.}}:
\bctitle{Co-evolutionary compression for unpaired image translation}.
In: \bbtitle{Proceedings of the IEEE/CVF International Conference on Computer
  Vision},
pp. \bfpage{3235}--\blpage{3244}
(\byear{2019})
\end{bchapter}
\endbibitem

\bibitem{imgtoimg_zhu2017unpaired}
\begin{bchapter}
\bauthor{\bsnm{Zhu}, \binits{J.-Y.}},
\bauthor{\bsnm{Park}, \binits{T.}},
\bauthor{\bsnm{Isola}, \binits{P.}},
\bauthor{\bsnm{Efros}, \binits{A.A.}}:
\bctitle{Unpaired image-to-image translation using cycle-consistent adversarial
  networks}.
In: \bbtitle{Proceedings of the IEEE International Conference on Computer
  Vision},
pp. \bfpage{2223}--\blpage{2232}
(\byear{2017})
\end{bchapter}
\endbibitem

\bibitem{SPGAN_song2020sp}
\begin{botherref}
\oauthor{\bsnm{Song}, \binits{X.}},
\oauthor{\bsnm{Chen}, \binits{Y.}},
\oauthor{\bsnm{Feng}, \binits{Z.-H.}},
\oauthor{\bsnm{Hu}, \binits{G.}},
\oauthor{\bsnm{Yu}, \binits{D.-J.}},
\oauthor{\bsnm{Wu}, \binits{X.-J.}}:
Sp-gan: Self-growing and pruning generative adversarial networks.
IEEE Transactions on Neural Networks and Learning Systems
(2020)
\end{botherref}
\endbibitem

\bibitem{GAN_COMPRESSION_DISTILLATION}
\begin{botherref}
\oauthor{\bsnm{Aguinaldo}, \binits{A.}},
\oauthor{\bsnm{Chiang}, \binits{P.-Y.}},
\oauthor{\bsnm{Gain}, \binits{A.}},
\oauthor{\bsnm{Patil}, \binits{A.}},
\oauthor{\bsnm{Pearson}, \binits{K.}},
\oauthor{\bsnm{Feizi}, \binits{S.}}:
Compressing gans using knowledge distillation.
arXiv preprint arXiv:1902.00159
(2019)
\end{botherref}
\endbibitem

\bibitem{KD_GAN_COMPRESSION_NAS}
\begin{bchapter}
\bauthor{\bsnm{Li}, \binits{M.}},
\bauthor{\bsnm{Lin}, \binits{J.}},
\bauthor{\bsnm{Ding}, \binits{Y.}},
\bauthor{\bsnm{Liu}, \binits{Z.}},
\bauthor{\bsnm{Zhu}, \binits{J.-Y.}},
\bauthor{\bsnm{Han}, \binits{S.}}:
\bctitle{Gan compression: Efficient architectures for interactive conditional
  gans}.
In: \bbtitle{Proceedings of the IEEE/CVF Conference on Computer Vision and
  Pattern Recognition},
pp. \bfpage{5284}--\blpage{5294}
(\byear{2020})
\end{bchapter}
\endbibitem

\bibitem{OFA_NAS}
\begin{botherref}
\oauthor{\bsnm{Cai}, \binits{H.}},
\oauthor{\bsnm{Gan}, \binits{C.}},
\oauthor{\bsnm{Wang}, \binits{T.}},
\oauthor{\bsnm{Zhang}, \binits{Z.}},
\oauthor{\bsnm{Han}, \binits{S.}}:
Once-for-all: Train one network and specialize it for efficient deployment.
arXiv preprint arXiv:1908.09791
(2019)
\end{botherref}
\endbibitem

\bibitem{KD_GAN_fu2020autogan}
\begin{botherref}
\oauthor{\bsnm{Fu}, \binits{Y.}},
\oauthor{\bsnm{Chen}, \binits{W.}},
\oauthor{\bsnm{Wang}, \binits{H.}},
\oauthor{\bsnm{Li}, \binits{H.}},
\oauthor{\bsnm{Lin}, \binits{Y.}},
\oauthor{\bsnm{Wang}, \binits{Z.}}:
Autogan-distiller: Searching to compress generative adversarial networks.
arXiv preprint arXiv:2006.08198
(2020)
\end{botherref}
\endbibitem

\bibitem{liu2018darts}
\begin{botherref}
\oauthor{\bsnm{Liu}, \binits{H.}},
\oauthor{\bsnm{Simonyan}, \binits{K.}},
\oauthor{\bsnm{Yang}, \binits{Y.}}:
Darts: Differentiable architecture search.
arXiv preprint arXiv:1806.09055
(2018)
\end{botherref}
\endbibitem

\bibitem{wang2018esrgan}
\begin{bchapter}
\bauthor{\bsnm{Wang}, \binits{X.}},
\bauthor{\bsnm{Yu}, \binits{K.}},
\bauthor{\bsnm{Wu}, \binits{S.}},
\bauthor{\bsnm{Gu}, \binits{J.}},
\bauthor{\bsnm{Liu}, \binits{Y.}},
\bauthor{\bsnm{Dong}, \binits{C.}},
\bauthor{\bsnm{Qiao}, \binits{Y.}},
\bauthor{\bsnm{Change~Loy}, \binits{C.}}:
\bctitle{Esrgan: Enhanced super-resolution generative adversarial networks}.
In: \bbtitle{Proceedings of the European Conference on Computer Vision (ECCV)
  Workshops},
pp. \bfpage{0}--\blpage{0}
(\byear{2018})
\end{bchapter}
\endbibitem

\bibitem{KD_GAN_CATjin2021teachers}
\begin{botherref}
\oauthor{\bsnm{Jin}, \binits{Q.}},
\oauthor{\bsnm{Ren}, \binits{J.}},
\oauthor{\bsnm{Woodford}, \binits{O.J.}},
\oauthor{\bsnm{Wang}, \binits{J.}},
\oauthor{\bsnm{Yuan}, \binits{G.}},
\oauthor{\bsnm{Wang}, \binits{Y.}},
\oauthor{\bsnm{Tulyakov}, \binits{S.}}:
Teachers do more than teach: Compressing image-to-image models.
arXiv preprint arXiv:2103.03467
(2021)
\end{botherref}
\endbibitem

\bibitem{KD_GAN_chen2020distilling}
\begin{bchapter}
\bauthor{\bsnm{Chen}, \binits{H.}},
\bauthor{\bsnm{Wang}, \binits{Y.}},
\bauthor{\bsnm{Shu}, \binits{H.}},
\bauthor{\bsnm{Wen}, \binits{C.}},
\bauthor{\bsnm{Xu}, \binits{C.}},
\bauthor{\bsnm{Shi}, \binits{B.}},
\bauthor{\bsnm{Xu}, \binits{C.}},
\bauthor{\bsnm{Xu}, \binits{C.}}:
\bctitle{Distilling portable generative adversarial networks for image
  translation}.
In: \bbtitle{Proceedings of the AAAI Conference on Artificial Intelligence},
vol. \bseriesno{34},
pp. \bfpage{3585}--\blpage{3592}
(\byear{2020})
\end{bchapter}
\endbibitem

\bibitem{KD_GAN_liu2021content}
\begin{botherref}
\oauthor{\bsnm{Liu}, \binits{Y.}},
\oauthor{\bsnm{Shu}, \binits{Z.}},
\oauthor{\bsnm{Li}, \binits{Y.}},
\oauthor{\bsnm{Lin}, \binits{Z.}},
\oauthor{\bsnm{Perazzi}, \binits{F.}},
\oauthor{\bsnm{Kung}, \binits{S.}}:
Content-aware gan compression.
arXiv preprint arXiv:2104.02244
(2021)
\end{botherref}
\endbibitem

\bibitem{KD_GAN_PKDGAN}
\begin{botherref}
\oauthor{\bsnm{Zhang}, \binits{Z.}},
\oauthor{\bsnm{Chen}, \binits{S.}},
\oauthor{\bsnm{Sun}, \binits{L.}}:
P-kdgan: Progressive knowledge distillation with gans for one-class novelty
  detection.
arXiv preprint arXiv:2007.06963
(2020)
\end{botherref}
\endbibitem

\bibitem{GAN_Slimming}
\begin{bchapter}
\bauthor{\bsnm{Wang}, \binits{H.}},
\bauthor{\bsnm{Gui}, \binits{S.}},
\bauthor{\bsnm{Yang}, \binits{H.}},
\bauthor{\bsnm{Liu}, \binits{J.}},
\bauthor{\bsnm{Wang}, \binits{Z.}}:
\bctitle{Gan slimming: All-in-one gan compression by a unified optimization
  framework}.
In: \bbtitle{European Conference on Computer Vision},
pp. \bfpage{54}--\blpage{73}
(\byear{2020}).
\bcomment{Springer}
\end{bchapter}
\endbibitem

\bibitem{liu2017learning}
\begin{bchapter}
\bauthor{\bsnm{Liu}, \binits{Z.}},
\bauthor{\bsnm{Li}, \binits{J.}},
\bauthor{\bsnm{Shen}, \binits{Z.}},
\bauthor{\bsnm{Huang}, \binits{G.}},
\bauthor{\bsnm{Yan}, \binits{S.}},
\bauthor{\bsnm{Zhang}, \binits{C.}}:
\bctitle{Learning efficient convolutional networks through network slimming}.
In: \bbtitle{Proceedings of the IEEE International Conference on Computer
  Vision},
pp. \bfpage{2736}--\blpage{2744}
(\byear{2017})
\end{bchapter}
\endbibitem

\bibitem{gogo_quantization}
\begin{bchapter}
\bauthor{\bsnm{Jacob}, \binits{B.}},
\bauthor{\bsnm{Kligys}, \binits{S.}},
\bauthor{\bsnm{Chen}, \binits{B.}},
\bauthor{\bsnm{Zhu}, \binits{M.}},
\bauthor{\bsnm{Tang}, \binits{M.}},
\bauthor{\bsnm{Howard}, \binits{A.}},
\bauthor{\bsnm{Adam}, \binits{H.}},
\bauthor{\bsnm{Kalenichenko}, \binits{D.}}:
\bctitle{Quantization and training of neural networks for efficient
  integer-arithmetic-only inference}.
In: \bbtitle{Proceedings of the IEEE Conference on Computer Vision and Pattern
  Recognition},
pp. \bfpage{2704}--\blpage{2713}
(\byear{2018})
\end{bchapter}
\endbibitem

\bibitem{QGAN}
\begin{botherref}
\oauthor{\bsnm{Wang}, \binits{P.}}:
"QGAN: Quantized Generative Adversarial Networks.".
"arXiv preprint".
(2019)
\end{botherref}
\endbibitem

\bibitem{WGAN-GP}
\begin{botherref}
\oauthor{\bsnm{Gulrajani}, \binits{I.}},
\oauthor{\bsnm{Ahmed}, \binits{F.}},
\oauthor{\bsnm{Arjovsky}, \binits{M.}},
\oauthor{\bsnm{Dumoulin}, \binits{V.}},
\oauthor{\bsnm{Courville}, \binits{A.}}:
Improved training of wasserstein gans.
arXiv preprint arXiv:1704.00028
(2017)
\end{botherref}
\endbibitem

\bibitem{LSGAN}
\begin{bchapter}
\bauthor{\bsnm{Mao}, \binits{X.}},
\bauthor{\bsnm{Li}, \binits{Q.}},
\bauthor{\bsnm{Xie}, \binits{H.}},
\bauthor{\bsnm{Lau}, \binits{R.Y.}},
\bauthor{\bsnm{Wang}, \binits{Z.}},
\bauthor{\bsnm{Paul~Smolley}, \binits{S.}}:
\bctitle{Least squares generative adversarial networks}.
In: \bbtitle{Proceedings of the IEEE International Conference on Computer
  Vision},
pp. \bfpage{2794}--\blpage{2802}
(\byear{2017})
\end{bchapter}
\endbibitem

\bibitem{QMGAN}
\begin{botherref}
\oauthor{\bsnm{Deng}, \binits{A.}},
\oauthor{\bsnm{Looi}, \binits{W.}},
\oauthor{\bsnm{Tsun}, \binits{A.}}:
Quantized GANs for Mobile Image Reconstruction
(2019)
\end{botherref}
\endbibitem

\bibitem{ApGAN_COM_PIM_DF_MEM}
\begin{botherref}
\oauthor{\bsnm{Roohi}, \binits{A.}},
\oauthor{\bsnm{Sheikhfaal}, \binits{S.}},
\oauthor{\bsnm{Angizi}, \binits{S.}},
\oauthor{\bsnm{Fan}, \binits{D.}},
\oauthor{\bsnm{DeMara}, \binits{R.F.}}:
Apgan: Approximate gan for robust low energy learning from imprecise
  components.
IEEE Transactions on Computers
(2019)
\end{botherref}
\endbibitem

\bibitem{PIMTGAN}
\begin{bchapter}
\bauthor{\bsnm{Rakin}, \binits{A.S.}},
\bauthor{\bsnm{Angizi}, \binits{S.}},
\bauthor{\bsnm{He}, \binits{Z.}},
\bauthor{\bsnm{Fan}, \binits{D.}}:
\bctitle{Pim-tgan: A processing-in-memory accelerator for ternary generative
  adversarial networks}.
In: \bbtitle{2018 IEEE 36th International Conference on Computer Design
  (ICCD)},
pp. \bfpage{266}--\blpage{273}
(\byear{2018}).
\bcomment{IEEE}
\end{bchapter}
\endbibitem

\bibitem{FlexPoint}
\begin{bchapter}
\bauthor{\bsnm{K{\"o}ster}, \binits{U.}},
\bauthor{\bsnm{Webb}, \binits{T.}},
\bauthor{\bsnm{Wang}, \binits{X.}},
\bauthor{\bsnm{Nassar}, \binits{M.}},
\bauthor{\bsnm{Bansal}, \binits{A.K.}},
\bauthor{\bsnm{Constable}, \binits{W.}},
\bauthor{\bsnm{Elibol}, \binits{O.}},
\bauthor{\bsnm{Gray}, \binits{S.}},
\bauthor{\bsnm{Hall}, \binits{S.}},
\bauthor{\bsnm{Hornof}, \binits{L.}}, \betal:
\bctitle{Flexpoint: An adaptive numerical format for efficient training of deep
  neural networks}.
In: \bbtitle{Advances in Neural Information Processing Systems},
pp. \bfpage{1742}--\blpage{1752}
(\byear{2017})
\end{bchapter}
\endbibitem

\bibitem{CNN_compression_Clustering}
\begin{botherref}
\oauthor{\bsnm{Han}, \binits{S.}},
\oauthor{\bsnm{Mao}, \binits{H.}},
\oauthor{\bsnm{Dally}, \binits{W.J.}}:
A deep neural network compression pipeline: Pruning, quantization, huffman
  encoding.
arXiv preprint arXiv:1510.00149
\textbf{10}
(2015)
\end{botherref}
\endbibitem

\bibitem{CNN_compression_resharing}
\begin{botherref}
\oauthor{\bsnm{Ullrich}, \binits{K.}},
\oauthor{\bsnm{Meeds}, \binits{E.}},
\oauthor{\bsnm{Welling}, \binits{M.}}:
Soft weight-sharing for neural network compression.
arXiv preprint arXiv:1702.04008
(2017)
\end{botherref}
\endbibitem

\bibitem{CNN_compression_hashing}
\begin{bchapter}
\bauthor{\bsnm{Chen}, \binits{W.}},
\bauthor{\bsnm{Wilson}, \binits{J.}},
\bauthor{\bsnm{Tyree}, \binits{S.}},
\bauthor{\bsnm{Weinberger}, \binits{K.}},
\bauthor{\bsnm{Chen}, \binits{Y.}}:
\bctitle{Compressing neural networks with the hashing trick}.
In: \bbtitle{International Conference on Machine Learning},
pp. \bfpage{2285}--\blpage{2294}
(\byear{2015})
\end{bchapter}
\endbibitem

\bibitem{CNN_compression_hashingBlock}
\begin{bchapter}
\bauthor{\bsnm{Zhu}, \binits{J.}},
\bauthor{\bsnm{Qian}, \binits{Z.}},
\bauthor{\bsnm{Tsui}, \binits{C.-Y.}}:
\bctitle{Bhnn: A memory-efficient accelerator for compressing deep neural
  networks with blocked hashing techniques}.
In: \bbtitle{2017 22nd Asia and South Pacific Design Automation Conference
  (ASP-DAC)},
pp. \bfpage{690}--\blpage{695}
(\byear{2017}).
\bcomment{IEEE}
\end{bchapter}
\endbibitem

\bibitem{CNN_compression_structuredhashing}
\begin{botherref}
\oauthor{\bsnm{Eban}, \binits{E.}},
\oauthor{\bsnm{Movshovitz-Attias}, \binits{Y.}},
\oauthor{\bsnm{Wu}, \binits{H.}},
\oauthor{\bsnm{Sandler}, \binits{M.}},
\oauthor{\bsnm{Poon}, \binits{A.}},
\oauthor{\bsnm{Idelbayev}, \binits{Y.}},
\oauthor{\bsnm{Carreira-Perpinan}, \binits{M.A.}}:
Structured multi-hashing for model compression.
arXiv preprint arXiv:1911.11177
(2019)
\end{botherref}
\endbibitem

\bibitem{RELATED_NOGAN_cheng2017survey}
\begin{botherref}
\oauthor{\bsnm{Cheng}, \binits{Y.}},
\oauthor{\bsnm{Wang}, \binits{D.}},
\oauthor{\bsnm{Zhou}, \binits{P.}},
\oauthor{\bsnm{Zhang}, \binits{T.}}:
A survey of model compression and acceleration for deep neural networks.
arXiv preprint arXiv:1710.09282
(2017)
\end{botherref}
\endbibitem

\bibitem{RELATED_NOGAN_cheng2018model}
\begin{barticle}
\bauthor{\bsnm{Cheng}, \binits{Y.}},
\bauthor{\bsnm{Wang}, \binits{D.}},
\bauthor{\bsnm{Zhou}, \binits{P.}},
\bauthor{\bsnm{Zhang}, \binits{T.}}:
\batitle{Model compression and acceleration for deep neural networks: The
  principles, progress, and challenges}.
\bjtitle{IEEE Signal Processing Magazine}
\bvolume{35}(\bissue{1}),
\bfpage{126}--\blpage{136}
(\byear{2018})
\end{barticle}
\endbibitem

\bibitem{RELATED_NOGAN_Cheng2018RecentAI}
\begin{barticle}
\bauthor{\bsnm{Cheng}, \binits{J.}},
\bauthor{\bsnm{Wang}, \binits{P.}},
\bauthor{\bsnm{Li}, \binits{G.}},
\bauthor{\bsnm{Hu}, \binits{Q.}},
\bauthor{\bsnm{Lu}, \binits{H.}}:
\batitle{Recent advances in efficient computation of deep convolutional neural
  networks}.
\bjtitle{Frontiers of Information Technology \& Electronic Engineering}
\bvolume{19},
\bfpage{64}--\blpage{77}
(\byear{2018})
\end{barticle}
\endbibitem

\bibitem{RELATED_NOGAN_choudhary2020comprehensive}
\begin{barticle}
\bauthor{\bsnm{Choudhary}, \binits{T.}},
\bauthor{\bsnm{Mishra}, \binits{V.}},
\bauthor{\bsnm{Goswami}, \binits{A.}},
\bauthor{\bsnm{Sarangapani}, \binits{J.}}:
\batitle{A comprehensive survey on model compression and acceleration}.
\bjtitle{Artificial Intelligence Review}
\bvolume{53}(\bissue{7}),
\bfpage{5113}--\blpage{5155}
(\byear{2020})
\end{barticle}
\endbibitem

\bibitem{RELATED_NOGAN_distill_usedGAN_gou2021knowledge}
\begin{barticle}
\bauthor{\bsnm{Gou}, \binits{J.}},
\bauthor{\bsnm{Yu}, \binits{B.}},
\bauthor{\bsnm{Maybank}, \binits{S.J.}},
\bauthor{\bsnm{Tao}, \binits{D.}}:
\batitle{Knowledge distillation: A survey}.
\bjtitle{International Journal of Computer Vision}
\bvolume{129}(\bissue{6}),
\bfpage{1789}--\blpage{1819}
(\byear{2021})
\end{barticle}
\endbibitem

\bibitem{RELATED_distill_usedGAN_Wang2021KnowledgeDA}
\begin{botherref}
\oauthor{\bsnm{Wang}, \binits{L.}},
\oauthor{\bsnm{Yoon}, \binits{K.-J.}}:
Knowledge distillation and student-teacher learning for visual intelligence: A
  review and new outlooks.
IEEE transactions on pattern analysis and machine intelligence
\textbf{PP}
(2021)
\end{botherref}
\endbibitem

\end{thebibliography}

\newcommand{\BMCxmlcomment}[1]{}

\BMCxmlcomment{

<refgrp>

<bibl id="B1">
  <title><p>Generative adversarial nets</p></title>
  <aug>
    <au><snm>Goodfellow</snm><fnm>I</fnm></au>
    <au><snm>Pouget Abadie</snm><fnm>J</fnm></au>
    <au><snm>Mirza</snm><fnm>M</fnm></au>
    <au><snm>Xu</snm><fnm>B</fnm></au>
    <au><snm>Warde Farley</snm><fnm>D</fnm></au>
    <au><snm>Ozair</snm><fnm>S</fnm></au>
    <au><snm>Courville</snm><fnm>A</fnm></au>
    <au><snm>Bengio</snm><fnm>Y</fnm></au>
  </aug>
  <source>Advances in neural information processing systems</source>
  <pubdate>2014</pubdate>
  <fpage>2672</fpage>
  <lpage>-2680</lpage>
</bibl>

<bibl id="B2">
  <title><p>Generative adversarial network-based glottal waveform model for
  statistical parametric speech synthesis</p></title>
  <aug>
    <au><snm>Bollepalli</snm><fnm>B</fnm></au>
    <au><snm>Juvela</snm><fnm>L</fnm></au>
    <au><snm>Alku</snm><fnm>P</fnm></au>
  </aug>
  <source>arXiv preprint arXiv:1903.05955</source>
  <pubdate>2019</pubdate>
</bibl>

<bibl id="B3">
  <title><p>Stackgan: Text to photo-realistic image synthesis with stacked
  generative adversarial networks</p></title>
  <aug>
    <au><snm>Zhang</snm><fnm>H</fnm></au>
    <au><snm>Xu</snm><fnm>T</fnm></au>
    <au><snm>Li</snm><fnm>H</fnm></au>
    <au><snm>Zhang</snm><fnm>S</fnm></au>
    <au><snm>Wang</snm><fnm>X</fnm></au>
    <au><snm>Huang</snm><fnm>X</fnm></au>
    <au><snm>Metaxas</snm><fnm>DN</fnm></au>
  </aug>
  <source>Proceedings of the IEEE international conference on computer
  vision</source>
  <pubdate>2017</pubdate>
  <fpage>5907</fpage>
  <lpage>-5915</lpage>
</bibl>

<bibl id="B4">
  <title><p>Photographic text-to-image synthesis with a hierarchically-nested
  adversarial network</p></title>
  <aug>
    <au><snm>Zhang</snm><fnm>Z</fnm></au>
    <au><snm>Xie</snm><fnm>Y</fnm></au>
    <au><snm>Yang</snm><fnm>L</fnm></au>
  </aug>
  <source>Proceedings of the IEEE Conference on Computer Vision and Pattern
  Recognition</source>
  <pubdate>2018</pubdate>
  <fpage>6199</fpage>
  <lpage>-6208</lpage>
</bibl>

<bibl id="B5">
  <title><p>Stargan: Unified generative adversarial networks for multi-domain
  image-to-image translation</p></title>
  <aug>
    <au><snm>Choi</snm><fnm>Y</fnm></au>
    <au><snm>Choi</snm><fnm>M</fnm></au>
    <au><snm>Kim</snm><fnm>M</fnm></au>
    <au><snm>Ha</snm><fnm>JW</fnm></au>
    <au><snm>Kim</snm><fnm>S</fnm></au>
    <au><snm>Choo</snm><fnm>J</fnm></au>
  </aug>
  <source>Proceedings of the IEEE conference on computer vision and pattern
  recognition</source>
  <pubdate>2018</pubdate>
  <fpage>8789</fpage>
  <lpage>-8797</lpage>
</bibl>

<bibl id="B6">
  <title><p>High-resolution image synthesis and semantic manipulation with
  conditional gans</p></title>
  <aug>
    <au><snm>Wang</snm><fnm>TC</fnm></au>
    <au><snm>Liu</snm><fnm>MY</fnm></au>
    <au><snm>Zhu</snm><fnm>JY</fnm></au>
    <au><snm>Tao</snm><fnm>A</fnm></au>
    <au><snm>Kautz</snm><fnm>J</fnm></au>
    <au><snm>Catanzaro</snm><fnm>B</fnm></au>
  </aug>
  <source>Proceedings of the IEEE conference on computer vision and pattern
  recognition</source>
  <pubdate>2018</pubdate>
  <fpage>8798</fpage>
  <lpage>-8807</lpage>
</bibl>

<bibl id="B7">
  <title><p>Gansynth: Adversarial neural audio synthesis</p></title>
  <aug>
    <au><snm>Engel</snm><fnm>J</fnm></au>
    <au><snm>Agrawal</snm><fnm>KK</fnm></au>
    <au><snm>Chen</snm><fnm>S</fnm></au>
    <au><snm>Gulrajani</snm><fnm>I</fnm></au>
    <au><snm>Donahue</snm><fnm>C</fnm></au>
    <au><snm>Roberts</snm><fnm>A</fnm></au>
  </aug>
  <source>arXiv preprint arXiv:1902.08710</source>
  <pubdate>2019</pubdate>
</bibl>

<bibl id="B8">
  <title><p>Efficient video generation on complex datasets</p></title>
  <aug>
    <au><snm>Clark</snm><fnm>A</fnm></au>
    <au><snm>Donahue</snm><fnm>J</fnm></au>
    <au><snm>Simonyan</snm><fnm>K</fnm></au>
  </aug>
  <source>arXiv preprint arXiv:1907.06571</source>
  <pubdate>2019</pubdate>
</bibl>

<bibl id="B9">
  <title><p>Fw-gan: Flow-navigated warping gan for video virtual
  try-on</p></title>
  <aug>
    <au><snm>Dong</snm><fnm>H</fnm></au>
    <au><snm>Liang</snm><fnm>X</fnm></au>
    <au><snm>Shen</snm><fnm>X</fnm></au>
    <au><snm>Wu</snm><fnm>B</fnm></au>
    <au><snm>Chen</snm><fnm>BC</fnm></au>
    <au><snm>Yin</snm><fnm>J</fnm></au>
  </aug>
  <source>Proceedings of the IEEE International Conference on Computer
  Vision</source>
  <pubdate>2019</pubdate>
  <fpage>1161</fpage>
  <lpage>-1170</lpage>
</bibl>

<bibl id="B10">
  <title><p>Fast Video Quality Enhancement using GANs</p></title>
  <aug>
    <au><snm>Galteri</snm><fnm>L</fnm></au>
    <au><snm>Seidenari</snm><fnm>L</fnm></au>
    <au><snm>Bertini</snm><fnm>M</fnm></au>
    <au><snm>Uricchio</snm><fnm>T</fnm></au>
    <au><snm>Del Bimbo</snm><fnm>A</fnm></au>
  </aug>
  <source>Proceedings of the 27th ACM International Conference on
  Multimedia</source>
  <pubdate>2019</pubdate>
  <fpage>1065</fpage>
  <lpage>-1067</lpage>
</bibl>

<bibl id="B11">
  <title><p>A deep convolutional generative adversarial networks (DCGANs)-based
  semi-supervised method for object recognition in synthetic aperture radar
  (SAR) images</p></title>
  <aug>
    <au><snm>Gao</snm><fnm>F</fnm></au>
    <au><snm>Yang</snm><fnm>Y</fnm></au>
    <au><snm>Wang</snm><fnm>J</fnm></au>
    <au><snm>Sun</snm><fnm>J</fnm></au>
    <au><snm>Yang</snm><fnm>E</fnm></au>
    <au><snm>Zhou</snm><fnm>H</fnm></au>
  </aug>
  <source>Remote Sensing</source>
  <publisher>Multidisciplinary Digital Publishing Institute</publisher>
  <pubdate>2018</pubdate>
  <volume>10</volume>
  <issue>6</issue>
  <fpage>846</fpage>
</bibl>

<bibl id="B12">
  <title><p>Image-to-image translation with conditional adversarial
  networks</p></title>
  <aug>
    <au><snm>Isola</snm><fnm>P</fnm></au>
    <au><snm>Zhu</snm><fnm>JY</fnm></au>
    <au><snm>Zhou</snm><fnm>T</fnm></au>
    <au><snm>Efros</snm><fnm>AA</fnm></au>
  </aug>
  <source>Proceedings of the IEEE conference on computer vision and pattern
  recognition</source>
  <pubdate>2017</pubdate>
  <fpage>1125</fpage>
  <lpage>-1134</lpage>
</bibl>

<bibl id="B13">
  <title><p>Improved techniques for training gans</p></title>
  <aug>
    <au><snm>Salimans</snm><fnm>T</fnm></au>
    <au><snm>Goodfellow</snm><fnm>I</fnm></au>
    <au><snm>Zaremba</snm><fnm>W</fnm></au>
    <au><snm>Cheung</snm><fnm>V</fnm></au>
    <au><snm>Radford</snm><fnm>A</fnm></au>
    <au><snm>Chen</snm><fnm>X</fnm></au>
  </aug>
  <source>Advances in neural information processing systems</source>
  <pubdate>2016</pubdate>
  <fpage>2234</fpage>
  <lpage>-2242</lpage>
</bibl>

<bibl id="B14">
  <title><p>Gans trained by a two time-scale update rule converge to a local
  nash equilibrium</p></title>
  <aug>
    <au><snm>Heusel</snm><fnm>M</fnm></au>
    <au><snm>Ramsauer</snm><fnm>H</fnm></au>
    <au><snm>Unterthiner</snm><fnm>T</fnm></au>
    <au><snm>Nessler</snm><fnm>B</fnm></au>
    <au><snm>Hochreiter</snm><fnm>S</fnm></au>
  </aug>
  <source>Advances in neural information processing systems</source>
  <pubdate>2017</pubdate>
  <fpage>6626</fpage>
  <lpage>-6637</lpage>
</bibl>

<bibl id="B15">
  <title><p>Self-Supervised GAN Compression</p></title>
  <aug>
    <au><snm>Yu</snm><fnm>C</fnm></au>
    <au><snm>Pool</snm><fnm>J</fnm></au>
  </aug>
  <source>arXiv:2007.01491v2</source>
  <pubdate>2020</pubdate>
</bibl>

<bibl id="B16">
  <title><p>Co-evolutionary compression for unpaired image
  translation</p></title>
  <aug>
    <au><snm>Shu</snm><fnm>H</fnm></au>
    <au><snm>Wang</snm><fnm>Y</fnm></au>
    <au><snm>Jia</snm><fnm>X</fnm></au>
    <au><snm>Han</snm><fnm>K</fnm></au>
    <au><snm>Chen</snm><fnm>H</fnm></au>
    <au><snm>Xu</snm><fnm>C</fnm></au>
    <au><snm>Tian</snm><fnm>Q</fnm></au>
    <au><snm>Xu</snm><fnm>C</fnm></au>
  </aug>
  <source>Proceedings of the IEEE/CVF International Conference on Computer
  Vision</source>
  <pubdate>2019</pubdate>
  <fpage>3235</fpage>
  <lpage>-3244</lpage>
</bibl>

<bibl id="B17">
  <title><p>Unpaired image-to-image translation using cycle-consistent
  adversarial networks</p></title>
  <aug>
    <au><snm>Zhu</snm><fnm>JY</fnm></au>
    <au><snm>Park</snm><fnm>T</fnm></au>
    <au><snm>Isola</snm><fnm>P</fnm></au>
    <au><snm>Efros</snm><fnm>AA</fnm></au>
  </aug>
  <source>Proceedings of the IEEE international conference on computer
  vision</source>
  <pubdate>2017</pubdate>
  <fpage>2223</fpage>
  <lpage>-2232</lpage>
</bibl>

<bibl id="B18">
  <title><p>SP-GAN: Self-growing and pruning generative adversarial
  networks</p></title>
  <aug>
    <au><snm>Song</snm><fnm>X</fnm></au>
    <au><snm>Chen</snm><fnm>Y</fnm></au>
    <au><snm>Feng</snm><fnm>ZH</fnm></au>
    <au><snm>Hu</snm><fnm>G</fnm></au>
    <au><snm>Yu</snm><fnm>DJ</fnm></au>
    <au><snm>Wu</snm><fnm>XJ</fnm></au>
  </aug>
  <source>IEEE Transactions on Neural Networks and Learning Systems</source>
  <publisher>IEEE</publisher>
  <pubdate>2020</pubdate>
</bibl>

<bibl id="B19">
  <title><p>Compressing gans using knowledge distillation</p></title>
  <aug>
    <au><snm>Aguinaldo</snm><fnm>A</fnm></au>
    <au><snm>Chiang</snm><fnm>PY</fnm></au>
    <au><snm>Gain</snm><fnm>A</fnm></au>
    <au><snm>Patil</snm><fnm>A</fnm></au>
    <au><snm>Pearson</snm><fnm>K</fnm></au>
    <au><snm>Feizi</snm><fnm>S</fnm></au>
  </aug>
  <source>arXiv preprint arXiv:1902.00159</source>
  <pubdate>2019</pubdate>
</bibl>

<bibl id="B20">
  <title><p>Gan compression: Efficient architectures for interactive
  conditional gans</p></title>
  <aug>
    <au><snm>Li</snm><fnm>M</fnm></au>
    <au><snm>Lin</snm><fnm>J</fnm></au>
    <au><snm>Ding</snm><fnm>Y</fnm></au>
    <au><snm>Liu</snm><fnm>Z</fnm></au>
    <au><snm>Zhu</snm><fnm>JY</fnm></au>
    <au><snm>Han</snm><fnm>S</fnm></au>
  </aug>
  <source>Proceedings of the IEEE/CVF Conference on Computer Vision and Pattern
  Recognition</source>
  <pubdate>2020</pubdate>
  <fpage>5284</fpage>
  <lpage>-5294</lpage>
</bibl>

<bibl id="B21">
  <title><p>Once-for-all: Train one network and specialize it for efficient
  deployment</p></title>
  <aug>
    <au><snm>Cai</snm><fnm>H</fnm></au>
    <au><snm>Gan</snm><fnm>C</fnm></au>
    <au><snm>Wang</snm><fnm>T</fnm></au>
    <au><snm>Zhang</snm><fnm>Z</fnm></au>
    <au><snm>Han</snm><fnm>S</fnm></au>
  </aug>
  <source>arXiv preprint arXiv:1908.09791</source>
  <pubdate>2019</pubdate>
</bibl>

<bibl id="B22">
  <title><p>Autogan-distiller: Searching to compress generative adversarial
  networks</p></title>
  <aug>
    <au><snm>Fu</snm><fnm>Y</fnm></au>
    <au><snm>Chen</snm><fnm>W</fnm></au>
    <au><snm>Wang</snm><fnm>H</fnm></au>
    <au><snm>Li</snm><fnm>H</fnm></au>
    <au><snm>Lin</snm><fnm>Y</fnm></au>
    <au><snm>Wang</snm><fnm>Z</fnm></au>
  </aug>
  <source>arXiv preprint arXiv:2006.08198</source>
  <pubdate>2020</pubdate>
</bibl>

<bibl id="B23">
  <title><p>Darts: Differentiable architecture search</p></title>
  <aug>
    <au><snm>Liu</snm><fnm>H</fnm></au>
    <au><snm>Simonyan</snm><fnm>K</fnm></au>
    <au><snm>Yang</snm><fnm>Y</fnm></au>
  </aug>
  <source>arXiv preprint arXiv:1806.09055</source>
  <pubdate>2018</pubdate>
</bibl>

<bibl id="B24">
  <title><p>Esrgan: Enhanced super-resolution generative adversarial
  networks</p></title>
  <aug>
    <au><snm>Wang</snm><fnm>X</fnm></au>
    <au><snm>Yu</snm><fnm>K</fnm></au>
    <au><snm>Wu</snm><fnm>S</fnm></au>
    <au><snm>Gu</snm><fnm>J</fnm></au>
    <au><snm>Liu</snm><fnm>Y</fnm></au>
    <au><snm>Dong</snm><fnm>C</fnm></au>
    <au><snm>Qiao</snm><fnm>Y</fnm></au>
    <au><snm>Change Loy</snm><fnm>C</fnm></au>
  </aug>
  <source>Proceedings of the European Conference on Computer Vision (ECCV)
  Workshops</source>
  <pubdate>2018</pubdate>
  <fpage>0</fpage>
  <lpage>-0</lpage>
</bibl>

<bibl id="B25">
  <title><p>Teachers Do More Than Teach: Compressing Image-to-Image
  Models</p></title>
  <aug>
    <au><snm>Jin</snm><fnm>Q</fnm></au>
    <au><snm>Ren</snm><fnm>J</fnm></au>
    <au><snm>Woodford</snm><fnm>OJ</fnm></au>
    <au><snm>Wang</snm><fnm>J</fnm></au>
    <au><snm>Yuan</snm><fnm>G</fnm></au>
    <au><snm>Wang</snm><fnm>Y</fnm></au>
    <au><snm>Tulyakov</snm><fnm>S</fnm></au>
  </aug>
  <source>arXiv preprint arXiv:2103.03467</source>
  <pubdate>2021</pubdate>
</bibl>

<bibl id="B26">
  <title><p>Distilling portable generative adversarial networks for image
  translation</p></title>
  <aug>
    <au><snm>Chen</snm><fnm>H</fnm></au>
    <au><snm>Wang</snm><fnm>Y</fnm></au>
    <au><snm>Shu</snm><fnm>H</fnm></au>
    <au><snm>Wen</snm><fnm>C</fnm></au>
    <au><snm>Xu</snm><fnm>C</fnm></au>
    <au><snm>Shi</snm><fnm>B</fnm></au>
    <au><snm>Xu</snm><fnm>C</fnm></au>
    <au><snm>Xu</snm><fnm>C</fnm></au>
  </aug>
  <source>Proceedings of the AAAI Conference on Artificial
  Intelligence</source>
  <pubdate>2020</pubdate>
  <volume>34</volume>
  <fpage>3585</fpage>
  <lpage>-3592</lpage>
</bibl>

<bibl id="B27">
  <title><p>Content-Aware GAN Compression</p></title>
  <aug>
    <au><snm>Liu</snm><fnm>Y</fnm></au>
    <au><snm>Shu</snm><fnm>Z</fnm></au>
    <au><snm>Li</snm><fnm>Y</fnm></au>
    <au><snm>Lin</snm><fnm>Z</fnm></au>
    <au><snm>Perazzi</snm><fnm>F</fnm></au>
    <au><snm>Kung</snm><fnm>SY</fnm></au>
  </aug>
  <source>arXiv preprint arXiv:2104.02244</source>
  <pubdate>2021</pubdate>
</bibl>

<bibl id="B28">
  <title><p>P-kdgan: Progressive knowledge distillation with gans for one-class
  novelty detection</p></title>
  <aug>
    <au><snm>Zhang</snm><fnm>Z</fnm></au>
    <au><snm>Chen</snm><fnm>S</fnm></au>
    <au><snm>Sun</snm><fnm>L</fnm></au>
  </aug>
  <source>arXiv preprint arXiv:2007.06963</source>
  <pubdate>2020</pubdate>
</bibl>

<bibl id="B29">
  <title><p>GAN Slimming: All-in-One GAN Compression by A Unified Optimization
  Framework</p></title>
  <aug>
    <au><snm>Wang</snm><fnm>H</fnm></au>
    <au><snm>Gui</snm><fnm>S</fnm></au>
    <au><snm>Yang</snm><fnm>H</fnm></au>
    <au><snm>Liu</snm><fnm>J</fnm></au>
    <au><snm>Wang</snm><fnm>Z</fnm></au>
  </aug>
  <source>European Conference on Computer Vision</source>
  <pubdate>2020</pubdate>
  <fpage>54</fpage>
  <lpage>-73</lpage>
</bibl>

<bibl id="B30">
  <title><p>Learning efficient convolutional networks through network
  slimming</p></title>
  <aug>
    <au><snm>Liu</snm><fnm>Z</fnm></au>
    <au><snm>Li</snm><fnm>J</fnm></au>
    <au><snm>Shen</snm><fnm>Z</fnm></au>
    <au><snm>Huang</snm><fnm>G</fnm></au>
    <au><snm>Yan</snm><fnm>S</fnm></au>
    <au><snm>Zhang</snm><fnm>C</fnm></au>
  </aug>
  <source>Proceedings of the IEEE international conference on computer
  vision</source>
  <pubdate>2017</pubdate>
  <fpage>2736</fpage>
  <lpage>-2744</lpage>
</bibl>

<bibl id="B31">
  <title><p>Quantization and training of neural networks for efficient
  integer-arithmetic-only inference</p></title>
  <aug>
    <au><snm>Jacob</snm><fnm>B</fnm></au>
    <au><snm>Kligys</snm><fnm>S</fnm></au>
    <au><snm>Chen</snm><fnm>B</fnm></au>
    <au><snm>Zhu</snm><fnm>M</fnm></au>
    <au><snm>Tang</snm><fnm>M</fnm></au>
    <au><snm>Howard</snm><fnm>A</fnm></au>
    <au><snm>Adam</snm><fnm>H</fnm></au>
    <au><snm>Kalenichenko</snm><fnm>D</fnm></au>
  </aug>
  <source>Proceedings of the IEEE conference on computer vision and pattern
  recognition</source>
  <pubdate>2018</pubdate>
  <fpage>2704</fpage>
  <lpage>-2713</lpage>
</bibl>

<bibl id="B32">
  <title><p>"QGAN: Quantized Generative Adversarial Networks."</p></title>
  <aug>
    <au><snm>Wang</snm><fnm>P</fnm></au>
  </aug>
  <publisher>"arXiv preprint".</publisher>
  <pubdate>2019</pubdate>
</bibl>

<bibl id="B33">
  <title><p>Improved training of wasserstein gans</p></title>
  <aug>
    <au><snm>Gulrajani</snm><fnm>I</fnm></au>
    <au><snm>Ahmed</snm><fnm>F</fnm></au>
    <au><snm>Arjovsky</snm><fnm>M</fnm></au>
    <au><snm>Dumoulin</snm><fnm>V</fnm></au>
    <au><snm>Courville</snm><fnm>A</fnm></au>
  </aug>
  <source>arXiv preprint arXiv:1704.00028</source>
  <pubdate>2017</pubdate>
</bibl>

<bibl id="B34">
  <title><p>Least squares generative adversarial networks</p></title>
  <aug>
    <au><snm>Mao</snm><fnm>X</fnm></au>
    <au><snm>Li</snm><fnm>Q</fnm></au>
    <au><snm>Xie</snm><fnm>H</fnm></au>
    <au><snm>Lau</snm><fnm>RY</fnm></au>
    <au><snm>Wang</snm><fnm>Z</fnm></au>
    <au><snm>Paul Smolley</snm><fnm>S</fnm></au>
  </aug>
  <source>Proceedings of the IEEE international conference on computer
  vision</source>
  <pubdate>2017</pubdate>
  <fpage>2794</fpage>
  <lpage>-2802</lpage>
</bibl>

<bibl id="B35">
  <title><p>Quantized GANs for Mobile Image Reconstruction</p></title>
  <aug>
    <au><snm>Deng</snm><fnm>A.</fnm></au>
    <au><snm>Looi</snm><fnm>W.</fnm></au>
    <au><snm>Tsun</snm><fnm>A.</fnm></au>
  </aug>
  <pubdate>2019</pubdate>
</bibl>

<bibl id="B36">
  <title><p>ApGAN: Approximate GAN for Robust Low Energy Learning from
  Imprecise Components</p></title>
  <aug>
    <au><snm>Roohi</snm><fnm>A</fnm></au>
    <au><snm>Sheikhfaal</snm><fnm>S</fnm></au>
    <au><snm>Angizi</snm><fnm>S</fnm></au>
    <au><snm>Fan</snm><fnm>D</fnm></au>
    <au><snm>DeMara</snm><fnm>RF</fnm></au>
  </aug>
  <source>IEEE Transactions on Computers</source>
  <publisher>IEEE</publisher>
  <pubdate>2019</pubdate>
</bibl>

<bibl id="B37">
  <title><p>PIM-TGAN: A processing-in-memory accelerator for ternary generative
  adversarial networks</p></title>
  <aug>
    <au><snm>Rakin</snm><fnm>AS</fnm></au>
    <au><snm>Angizi</snm><fnm>S</fnm></au>
    <au><snm>He</snm><fnm>Z</fnm></au>
    <au><snm>Fan</snm><fnm>D</fnm></au>
  </aug>
  <source>2018 IEEE 36th International Conference on Computer Design
  (ICCD)</source>
  <pubdate>2018</pubdate>
  <fpage>266</fpage>
  <lpage>-273</lpage>
</bibl>

<bibl id="B38">
  <title><p>Flexpoint: An adaptive numerical format for efficient training of
  deep neural networks</p></title>
  <aug>
    <au><snm>K{\"o}ster</snm><fnm>U</fnm></au>
    <au><snm>Webb</snm><fnm>T</fnm></au>
    <au><snm>Wang</snm><fnm>X</fnm></au>
    <au><snm>Nassar</snm><fnm>M</fnm></au>
    <au><snm>Bansal</snm><fnm>AK</fnm></au>
    <au><snm>Constable</snm><fnm>W</fnm></au>
    <au><snm>Elibol</snm><fnm>O</fnm></au>
    <au><snm>Gray</snm><fnm>S</fnm></au>
    <au><snm>Hall</snm><fnm>S</fnm></au>
    <au><snm>Hornof</snm><fnm>L</fnm></au>
    <au><cnm>others</cnm></au>
  </aug>
  <source>Advances in neural information processing systems</source>
  <pubdate>2017</pubdate>
  <fpage>1742</fpage>
  <lpage>-1752</lpage>
</bibl>

<bibl id="B39">
  <title><p>A deep neural network compression pipeline: Pruning, quantization,
  huffman encoding</p></title>
  <aug>
    <au><snm>Han</snm><fnm>S</fnm></au>
    <au><snm>Mao</snm><fnm>H</fnm></au>
    <au><snm>Dally</snm><fnm>WJ</fnm></au>
  </aug>
  <source>arXiv preprint arXiv:1510.00149</source>
  <pubdate>2015</pubdate>
  <volume>10</volume>
</bibl>

<bibl id="B40">
  <title><p>Soft weight-sharing for neural network compression</p></title>
  <aug>
    <au><snm>Ullrich</snm><fnm>K</fnm></au>
    <au><snm>Meeds</snm><fnm>E</fnm></au>
    <au><snm>Welling</snm><fnm>M</fnm></au>
  </aug>
  <source>arXiv preprint arXiv:1702.04008</source>
  <pubdate>2017</pubdate>
</bibl>

<bibl id="B41">
  <title><p>Compressing neural networks with the hashing trick</p></title>
  <aug>
    <au><snm>Chen</snm><fnm>W</fnm></au>
    <au><snm>Wilson</snm><fnm>J</fnm></au>
    <au><snm>Tyree</snm><fnm>S</fnm></au>
    <au><snm>Weinberger</snm><fnm>K</fnm></au>
    <au><snm>Chen</snm><fnm>Y</fnm></au>
  </aug>
  <source>International conference on machine learning</source>
  <pubdate>2015</pubdate>
  <fpage>2285</fpage>
  <lpage>-2294</lpage>
</bibl>

<bibl id="B42">
  <title><p>BHNN: A memory-efficient accelerator for compressing deep neural
  networks with blocked hashing techniques</p></title>
  <aug>
    <au><snm>Zhu</snm><fnm>J</fnm></au>
    <au><snm>Qian</snm><fnm>Z</fnm></au>
    <au><snm>Tsui</snm><fnm>CY</fnm></au>
  </aug>
  <source>2017 22nd Asia and South Pacific Design Automation Conference
  (ASP-DAC)</source>
  <pubdate>2017</pubdate>
  <fpage>690</fpage>
  <lpage>-695</lpage>
</bibl>

<bibl id="B43">
  <title><p>Structured Multi-Hashing for Model Compression</p></title>
  <aug>
    <au><snm>Eban</snm><fnm>E</fnm></au>
    <au><snm>Movshovitz Attias</snm><fnm>Y</fnm></au>
    <au><snm>Wu</snm><fnm>H</fnm></au>
    <au><snm>Sandler</snm><fnm>M</fnm></au>
    <au><snm>Poon</snm><fnm>A</fnm></au>
    <au><snm>Idelbayev</snm><fnm>Y</fnm></au>
    <au><snm>Carreira Perpinan</snm><fnm>MA</fnm></au>
  </aug>
  <source>arXiv preprint arXiv:1911.11177</source>
  <pubdate>2019</pubdate>
</bibl>

<bibl id="B44">
  <title><p>A survey of model compression and acceleration for deep neural
  networks</p></title>
  <aug>
    <au><snm>Cheng</snm><fnm>Y</fnm></au>
    <au><snm>Wang</snm><fnm>D</fnm></au>
    <au><snm>Zhou</snm><fnm>P</fnm></au>
    <au><snm>Zhang</snm><fnm>T</fnm></au>
  </aug>
  <source>arXiv preprint arXiv:1710.09282</source>
  <pubdate>2017</pubdate>
</bibl>

<bibl id="B45">
  <title><p>Model compression and acceleration for deep neural networks: The
  principles, progress, and challenges</p></title>
  <aug>
    <au><snm>Cheng</snm><fnm>Y</fnm></au>
    <au><snm>Wang</snm><fnm>D</fnm></au>
    <au><snm>Zhou</snm><fnm>P</fnm></au>
    <au><snm>Zhang</snm><fnm>T</fnm></au>
  </aug>
  <source>IEEE Signal Processing Magazine</source>
  <publisher>IEEE</publisher>
  <pubdate>2018</pubdate>
  <volume>35</volume>
  <issue>1</issue>
  <fpage>126</fpage>
  <lpage>-136</lpage>
</bibl>

<bibl id="B46">
  <title><p>Recent advances in efficient computation of deep convolutional
  neural networks</p></title>
  <aug>
    <au><snm>Cheng</snm><fnm>J</fnm></au>
    <au><snm>Wang</snm><fnm>P</fnm></au>
    <au><snm>Li</snm><fnm>G.</fnm></au>
    <au><snm>Hu</snm><fnm>Q</fnm></au>
    <au><snm>Lu</snm><fnm>H</fnm></au>
  </aug>
  <source>Frontiers of Information Technology \& Electronic
  Engineering</source>
  <pubdate>2018</pubdate>
  <volume>19</volume>
  <fpage>64</fpage>
  <lpage>77</lpage>
</bibl>

<bibl id="B47">
  <title><p>A comprehensive survey on model compression and
  acceleration</p></title>
  <aug>
    <au><snm>Choudhary</snm><fnm>T</fnm></au>
    <au><snm>Mishra</snm><fnm>V</fnm></au>
    <au><snm>Goswami</snm><fnm>A</fnm></au>
    <au><snm>Sarangapani</snm><fnm>J</fnm></au>
  </aug>
  <source>Artificial Intelligence Review</source>
  <publisher>Springer</publisher>
  <pubdate>2020</pubdate>
  <volume>53</volume>
  <issue>7</issue>
  <fpage>5113</fpage>
  <lpage>-5155</lpage>
</bibl>

<bibl id="B48">
  <title><p>Knowledge distillation: A survey</p></title>
  <aug>
    <au><snm>Gou</snm><fnm>J</fnm></au>
    <au><snm>Yu</snm><fnm>B</fnm></au>
    <au><snm>Maybank</snm><fnm>SJ</fnm></au>
    <au><snm>Tao</snm><fnm>D</fnm></au>
  </aug>
  <source>International Journal of Computer Vision</source>
  <publisher>Springer</publisher>
  <pubdate>2021</pubdate>
  <volume>129</volume>
  <issue>6</issue>
  <fpage>1789</fpage>
  <lpage>-1819</lpage>
</bibl>

<bibl id="B49">
  <title><p>Knowledge Distillation and Student-Teacher Learning for Visual
  Intelligence: A Review and New Outlooks</p></title>
  <aug>
    <au><snm>Wang</snm><fnm>L</fnm></au>
    <au><snm>Yoon</snm><fnm>KJ</fnm></au>
  </aug>
  <source>IEEE transactions on pattern analysis and machine
  intelligence</source>
  <pubdate>2021</pubdate>
  <volume>PP</volume>
</bibl>

</refgrp>
} 



\end{backmatter}
\end{document}